\documentclass{article}

\usepackage{arxiv}

\usepackage[utf8]{inputenc} 
\usepackage[T1]{fontenc}    
\usepackage{hyperref}       
\usepackage{url}            
\usepackage{booktabs}       
\usepackage{amsfonts}       
\usepackage{nicefrac}       
\usepackage{microtype}      
\usepackage{lipsum}		
\usepackage{graphicx}
\usepackage{doi}
\usepackage{adjustbox}
\usepackage{amsmath}
\usepackage{algpseudocode}
\usepackage{algorithm}
\usepackage{mfirstuc}
\usepackage{mathtools}
\usepackage{stackrel}
\usepackage{array}
\usepackage{xcolor,colortbl}

\definecolor{Gray}{gray}{0.85}
\definecolor{LightCyan}{rgb}{0.88,1,1}

\newcolumntype{a}{>{\columncolor{Gray}}c}
\newcolumntype{b}{>{\columncolor{white}}c}

\newcolumntype{P}[1]{>{\centering\arraybackslash}p{#1}}
\algnewcommand\Andi{\textbf{and}}
\algnewcommand\Or{\textbf{or}}

\title{Online Multi-Object Tracking with $\delta$-GLMB Filter based on Occlusion and Identity Switch Handling}

\author{Mohammadjavad Abbaspour\\
	School of Electrical and Computer Engineering\\
	Shiraz University\\
	Shiraz, Iran \\
	\texttt{mj.abbaspour@shirazu.ac.ir} \\
	\And
	Mohammad Ali Masnadi-Shirazi \\
	School of Electrical and Computer Engineering\\
	Shiraz University\\
	Shiraz, Iran \\
	\texttt{masnadi@shirazu.ac.ir} \\
}

\date{}

\renewcommand{\headeright}{}
\renewcommand{\undertitle}{}
\renewcommand{\shorttitle}{}

\begin{document}
\maketitle

\begin{abstract}
In this paper, we propose an online multi-object tracking (MOT) method in a delta Generalized Labeled Multi-Bernoulli ($\delta$-GLMB) filter framework to address occlusion and miss-detection issues, reduce false alarms, and recover identity switch (ID switch). To handle occlusion and miss-detection issues, we propose a measurement-to-disappeared track association method based on one-step $\delta$-GLMB filter, so it is possible to manage these difficulties by jointly processing occluded or miss-detected objects. The proposed method is based on a proposed similarity metric which is responsible for defining the weight of hypothesized reappeared tracks. We also extend the $\delta$-GLMB filter to efficiently recover switched IDs using the cardinality density, size and color features of the hypothesized tracks. We also propose a novel birth model to achieve more effective clutter removal performance. In both occlusion/miss-detection handler and newly-birthed object detector sections of the proposed method, unassigned measurements play a significant role, since they are used as the candidates for reappeared or birth objects. We evaluate the proposed method on well-known and publicly available MOT15 and MOT17 test datasets which are focused on pedestrian tracking. Experimental results show that the proposed tracker performs better or at least at the same level of the state-of-the-art online and offline MOT methods. It effectively handles the occlusion and ID switch issues and reduces false alarms as well.
\end{abstract}


\section{Introduction}
\label{sec:1}

Visual multi-object tracking has been a major topic in machine vision in the recent years. Its diverse applications such as autonomous driving and video surveillance has made visual MOT a necessary task and highly addressed problem. In literature, many methods are proposed to address this problem in order to solve its challenging issues \cite{song2019online, ong2020bayesian, kim2019labeled, rathnayake2020line}. Estimating the state of variable number of objects, continuously tracking objects over time and addressing occlusion and miss-detection are the main challenges of MOT.\\
A common approach to perform visual MOT task is tracking-by-detection \cite{wu2007detection, breitenstein2010online, yoon2015bayesian, fu2019multi}. Methods, which apply this approach, are based on using detections as the inputs of the tracking module. Motion analysis using a set of detections that represent an image, is a key technique used in this approach. Although this representation leads to information loss, it has increased the method's utility. On the other hand, track-before-detect (TBD) is a methodology that extracts spatio-temporal information from the image and not from the detections \cite{abbaspour2014robust, davey2012track, isard2001bramble, papi2015generalized}. The advantage of TBD approach is its prevention of information loss, while it is at the cost of accepting more computational load which has made it unpopular in real-time applications.\\
Visual MOT methods which exploit tracking-by-detection approach can be accomplished in two ways: online and offline. In online MOT, it is possible to use only current and past measurements to estimate the current states of the objects \cite{fu2018particle, xiang2015learning, bae2017confidence, yang2017hybrid, yang2016temporal}, while in offline MOT, whole measurements (past, current and future) are available to perform state estimation in each time \cite{xiang2020end, peng2020tpm, keuper2018motion, zhang2020long}. Online methods are faster because of less involved data, while offline methods are more accurate. In most of applications, such as autonomous vehicle systems \cite{wang2017pedestrian, sun2019model}, it is essential to perform the MOT task online. Due to the fact that future measurements cannot be utilized to address problems in online approaches, proposed methods are concentrated on somehow predicting the states of disappeared (miss-detected) tracks. In the view of handling MOT issues, methods that are based on Random Finite Sets (RFS) theory \cite{song2019online, fu2019multi, feng2016social, dames2019distributed, granstrom2012extended} have shown better or at least the same level of performance in comparison with other state-of-the-art online methods, since they are completely adopted for MOT task. Considering birth and death of objects, miss-detections and clutters followed by participating label of objects in its framework, allow to handle challenging issues of MOT straightforwardly.  \\
In this work, we apply online detection-based tracking while using visual features in specific situations in order to improve performance. Poor implementation of tracking submodules leads to severe performance drop in MOT task. However, drawbacks of the detection module are the main reason of facing challenging issues such as ID switch, track discontinuity, occlusion and miss-detection. As a general rule, using detections which are more trusted, results in a better final performance. So applying first-rate detectors such as FRCNN \cite{ren2015faster} and YOLO \cite{redmon2016you} leads to better tracking performance. Thanks to considerable progress in detectors' performance, false positive (clutter) is a rare issue in MOT, while because of the nature of the problem, false negative (miss-detection/occlusion) frequently happens. \\
In the last decade, many state-of-the-art methods have been proposed to tackle the noted challenges. Despite recent improvements, online MOT task is still an open area of research. Hence, this paper proposes a novel MOT method to address the existing challenges. Our method is based on an expansion of $\delta$-GLMB filter. So our main contributions are described as follows:\\
1) Providing a new method of detecting newly-birthed objects in which we confront false positive birth candidates. Although false positives are rarely produced by detection module, data association driven in birth process creates some amount of false positives. So we propose a new birth model based on $\delta$-GLMB filter which results in reducing false positives and removing clutters. \\
2) Expanding the $\delta$-GLMB filter in order to tackle the ID switch problem. Track consistency is a major requirement in any visual MOT application. In this work, we propose a new method based on $\delta$-GLMB filter to handle this issue. We use visual features in order to detect ID switches. So we prune out hypotheses that include ID switched objects and repeat state estimation task to achieve acceptable results according to a specific rule that we will describe later.\\
3) Proposing an expansion of $\delta$-GLMB filter that strongly addresses the occlusion and miss-detection problems in separated but identical ways. We develop one-step time $\delta$-GLMB filters which are devised to handle occlusion and miss-detection efficiently based on Bayesian inference. In this work, we use unassigned measurements as reappeared objects' candidates which are evaluated by incorporating a weighting system based on a new distance metric. \\
The rest of the paper is organized as follows: The related works are described in section \ref{sec:2}. In section \ref{sec:3}, the proposed method is elaborated, followed by performance evaluation of the method against other state-of-the-art methods in section \ref{sec:4}. Finally we conclude this paper in section \ref{sec:5}.

\section{Related Works}	
\label{sec:2}
We briefly introduced different types of MOT methods in the section \ref{sec:1}. Recently, TBD algorithms utilize first-rate detectors such as FRCNN \cite{ren2015faster} and YOLO \cite{redmon2016you}, which are based on deep neural networks (DNN). Although they have shown acceptable performance in multi-object detection task, due to illumination changes, miss-detection and occlusion issues they are imperfect in performing data association on their own. So many proposed algorithms utilize motion features along with visual features. \cite{yoon2018online} proposed historical appearance matching algorithm and adaptively defined detection confidence threshold in order to overcome problems created by false positives and false negatives returned by a DNN based detector. \cite{chen2018Ai} proposed a DNN based score function in order to perform optimal selection between detections and predicted tracks returned by motion analysis part. The method also includes a DNN based re-identification system to perform appearance matching.\cite{xiang2020end} utilized end-to-end learning framework to learn a deep conditional random field network in order to tackle occlusion and ID switch issues. Their two-step training strategy is based on considering unary and pairwise cases along with using long short-term memory (LSTM) to efficiently handle exploiting dependencies for pairwise terms. In \cite{zhang2020long} an iterative clustering method is proposed in order to generate tracklets which are introduced as an effective way to overcome numerous ID switches in MOT task. A deep association method based on LSTM networks is introduced which associates tracklets based on motion and appearance features. Two separate networks are used to perform learning each type of features in order to associate tracklets and generate complete trajectories.\\
On the other hand, detection-based methods mostly rely on the motion features and not the visual ones. This type of methods work by picking the best path based on the past and present measurement sets consisting of noisy position of object(s) and clutter. It becomes an effortful process when the number of objects is uncertain and time-varying. In recent years, RFS theory has been utilized frequently to address MOT task \cite{song2019online, fu2019multi, feng2016social, dames2019distributed, granstrom2012extended}. Using RFS, the objects' states, the number of objects and their identity are integrated and treated as a random set. Defining MOT problem in this mathematical framework allows to estimate the number of objects and their states which are ordinary tasks of a tracking method. In addition to considering birth and death of objects, RFS tracking methods also allow us to incorporate clutter and missed detections (occlusions) in the mathematical framework. Mahler presented Finite set statistics (FISST) \cite{mahler1994global} and \cite{mahler1994random} as a mathematical tool to address the calculus and statistics of RFSs, resulting in derivation of multi-object Bayesian filter, a non-trivial extension of the single object Bayesian filter. The multi-object Bayesian filter generally is computationally intractable. Mahler, then, presented the first-moment approximation of the intractable multi-object Bayesian filter and dubbed it as probability hypothesis density (PHD) filter \cite{mahler2003multitarget}. Mahler also introduced cardinality PHD (CPHD) filter, which incorporates estimation of number of objects by propagating cardinality distribution \cite{mahler2006theory}. Although CPHD filter outperforms PHD filter, it is followed by increasing in computational load. PHD and CPHD filters do not address identity estimation and extra considerations is needed. Recently, a new RFS based MOT approach, Generalized Labeled Multi-Bernoulli (GLMB) filter \cite{vo2013labeled}, is introduced which includes the identity of the objects. In contrast with other RFS based filters, GLMB filter does not need to apply approximation to be tractable, while it is an efficient framework driven based on Bayesian inference.  In the view of accuracy, GLMB filter outperforms other RFS based filters at the cost of more computational load. Remedies to reduce computational costs have been presented in \cite{vo2014labeled} and \cite{vo2016efficient}. In this work we apply $\delta$-GLMB filter \cite{vo2016efficient} to effectively address MOT issues. Note that one of our contributions is to reduce hypotheses in MOT task in order to decrease computational costs. \\
The literature review shows that there has been a gradual increase in the use of RFS based filters in proposed MOT methods due to advantages we discussed earlier. \cite{fu2017enhanced} proposed an enhanced Gaussian mixture PHD (GM-PHD) filter in which a classification step is applied to distinguish false measurements using confidence score and a gating technique in order to improve tracking performance in crowded scenes. A pre-trained convolutional neural network (CNN) is also applied to extract human features used in their proposed track management method to overcome occlusion issue. \cite{baisa2019occlusion} applied GM-PHD filter and CNN representations learning in order to devise an occlusion-robust visual tracker. An adaptive threshold is defined to select confident detection for birth tracks in order to reduce false positives (clutter). Furthermore, the proposed method considers unassigned track predictions to overcome occlusion. \cite{rathnayake2020line} developed a visual multi-object tracker which applies GLMB filter in order to overcome occlusion and track discontinuity issues. A likelihood function, constructed based on appearance and motion features, is proposed which defines whether or not a birth track is a reappeared one. To do this, they applied an organized lookup table for disappeared tracks. \cite{kim2017online} also selected GLMB filter paradigm and used appearance features returned form a pre-trained CNN to perform online MOT. \cite{baisa2019development} developed an extension of GM-PHD filter named N-type PHD filter to perform multiple object tracking by taking in to account multiple types of objects. Accordingly their method is able to deal with confused detections returning by multiple detectors. \cite{kim2019labeled} proposed a new measurement model for GLMB filter which considers both detections and image observations. The expanded GLMB filter distributions are shown to include conjugacy property with respect to new measurement model. \cite{song2019online} developed GM-PHD filter according to a new measurement-to-track and track-to-track association cost functions. They also introduced a new energy function to minimize occlusion group energy in order to handle the occlusion issue. \cite{fu2019multi} utilized GM-PHD filter along with a multi-level cooperative fusion approach to handle the MOT task. Full-body and body-parts detections are provided by two detectors to effectively address track-to-track association. Two separate discriminative correlation filters are proposed to effectively discriminate between objects and background, so objects' appearances are encoded and fused with spatio-temporal information in order to reduce wrong identity allocation. Finally a fusion center is devised to perform track management and track fusion using Generalized Correlation Intersection (GCI) rule.\\
Different from the above works, we use $\delta$-GLMB filter to tackle MOT issues not by adding different modules to the baseline filter but by expanding the baseline filter in order to equip it based on addressing each issue. So we can benefit from the efficiency of the filter and its MOT-compatible nature. 

\section{Proposed Method}
\label{sec:3}
\begin{figure*}
	\centering
		\scalebox{0.5}{\includegraphics[width=1\linewidth]{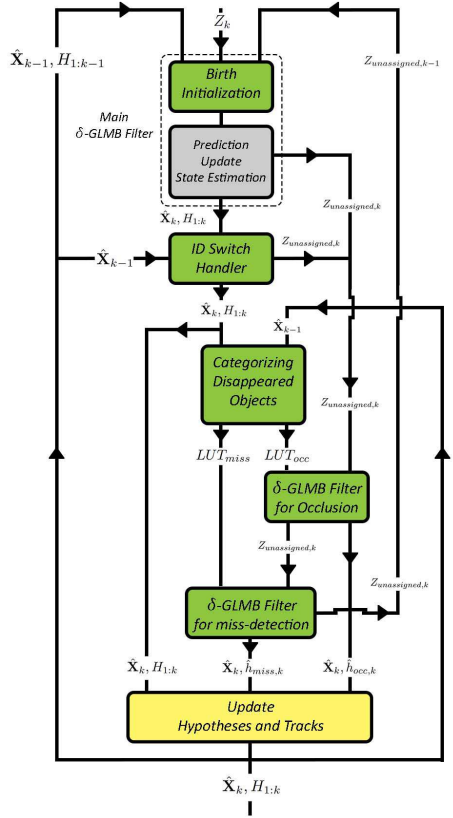}}

	\caption{The overall flow chart of the proposed method. The dotted rectangular contains the main $\delta$-GLMB filter. Major components are introduced in subsections \ref{sec:3_1} to \ref{sec:3_4}.}
	\label{fig:1}
\end{figure*}

In this section, we introduce the $\delta$-GLMB filter in subsection \ref{sec:3_1}. In subsections \ref{sec:3_2}, \ref{sec:3_3} and \ref{sec:3_4}, we present our proposed method including expanded $\delta$-GLMB filter to address the existing MOT issues. Figure \ref{fig:1} illustrates overall flowchart of our proposed method. In subsection \ref{sec:3_2}, we present our proposed birth model as a solution for detecting newly birthed objects without producing wrong birth candidates (false positive tracks). In subsection \ref{sec:3_3}, we introduce our method to tackle the problem of ID switch in order to achieve track continuity. Finally, We introduce our proposed method to address miss-detection and occlusion problems in subsection \ref{sec:3_4}. \\

\subsection{GLMB Filter}
\label{sec:3_1}
As we mentioned earlier, in MOT, there exists uncertainty about the number of objects, because of a couple of probable births and deaths in each time step and unfavorable situations such as occlusion and miss-detection. Accordingly, trajectory consistency becomes a major problem and it is vital to be addressed. In this paper we extend the GLMB filter, a (FISST)-based filter, proposed in \cite{vo2016efficient} to tackle the problem. The main reason we choose to use GLMB filter is that it remarkably fits to the MOT issue. In GLMB filter, MOT is well modeled by formulating a group of objects as a random finite set and can be extended to perform visual MOT and address the related problems.\\
The main advantage of GLMB filter over the other (FISST)-based filters such as PHD and CPHD is that the labels of the objects are included in the formulations. So the GLMB filter propagates the labeled set densities. In the following, the mathematical formulation of this filter along with used notations are introduced.\\
In this paper, single-object states are represented by lower-case letters (e.g. $\textbf{x}$ and $x$) while multi-object states are denoted by upper-case letters (e.g. $\textbf{X}$ and $X$). We also use bolded letters for labeled states, while unlabeled ones are not bolded. Furthermore spaces are denoted by blackboard letters (e.g. $\mathbb{L}$, $\mathbb{X}$, $\mathbb{B}$ and $\mathbb{N}$). The generalized Kroner delta is denoted by $\delta_{Y}[X]$ which is formulated as below:\\
\begin{equation}
	\label{eq:1}
	\delta_{Y}[X]=
	\begin{cases}
		1, & X = Y\\
		0, & otherwise\\
	\end{cases}
\end{equation}
The input argument takes any variable type such as set, vector or scalar. Consider $S$ as a set, $1_S(X)$, the indicator function, is given as below:\\
\begin{equation}
	\label{eq:2}
	1_S(X)=
	\begin{cases}
		1, & X \in S\\
		0, & otherwise\\
	\end{cases}
\end{equation}
In FISST, the product $\prod_{\substack{x \in X}}\; f(x)$ is denoted by $f^X$ and $f^\emptyset =1$. Throughout the paper, we use the notation ${\langle f, g \rangle}$ for the inner product $\int f(x)g(x) dx$. 
A labeled random finite set variable $(x,\ell)$ with the state $x \in \mathbb{X}$ (state space) and the label $\ell \in \mathbb{L}$ (a discrete space), is a subset of the space $\mathbb{X} \times \mathbb{L}$. The projection $\mathcal{L}:\mathbb{X} \times \mathbb{L} \rightarrow \mathbb{L}$ gives the label of a labeled RFS, e.g., $\mathcal{L}((x_1,\ell_1),(x_2,\ell_2))=\{\ell_1,\ell_2\}$. Note that a labeled RFS process is a point process such that each realization has distinct labels, so $\textbf{X}$ and $\mathcal{L}(\textbf{X})$ have the same cardinality. Accordingly, $\Delta(\textbf{X})$ is an indictor to show that whether or not labeled RFS members have distinct labels and is given as below:
\begin{equation}
	\label{eq:3}
	\Delta(X) = \delta_{|\textbf{X}|}[|\mathcal{L}(\textbf{X})|]
\end{equation}
A GLMB density is defined as follows:
\begin{equation}
	\label{eq:4}
	\boldsymbol{\pi}(\textbf{X})=\Delta(\textbf{X})
	\sum_{\substack{\xi \in \Xi}} \omega^{(\xi)}(\mathcal{L}(\textbf{X}))\Big{[}p^{(\xi)}\Big{]}^{\textbf{X}}
\end{equation}
where $\Xi$ is a discrete space and $\omega^{(\xi)}(L)$ is a non-negative weight associated with discrete index $\xi$ and $\textbf{X}$ with $\mathcal{L}(\textbf{X}) = L$,  while $\sum_{\substack{\xi \in \Xi}} \sum_{\substack{L \in \mathcal{F}(\mathbb{L})}}\omega^{(\xi)}(L) = 1$. Moreover, $p^{(\xi)}$ is a probability density function on $\mathbb{X}$ and connected with $\xi$.\\
It is shown that GLMB density function is closed under Bayesian inference for the standard multi-object transition and measurement models \cite{vo2013labeled}. So the recursive nature of Bayesian filtering keeps unchanged the mathematical form of the densities. The recently introduced compatible form of GLMB filter for multi-object tracking, $\delta$-GLMB filter, is also proven to be close under Bayesian inference. It is formulated as follows:\\  
\begin{equation}
	\label{eq:5}
	\boldsymbol{\pi}(\textbf{X})=\Delta(\textbf{X})
	\sum_{\substack{\xi \in \Xi}} \sum_{\substack{\substack{I \in \mathcal{F}(\mathbb{L})}}} \omega^{(I,\xi)}\delta_I\big{[}\mathcal{L}(\textbf{X})\big{]} \Big{[}p^{(\xi)}\Big{]}^{\textbf{X}}
\end{equation}
where $\Xi$ is the set of association maps history and $\mathcal{F}(\mathbb{L})$ is a finite subset of $\mathbb{L}$ , a discrete space including object labels. Each pair of $I$ and $\xi$ makes a component of the density which is the result of a track to measurement association and is described by $\omega^{(I,\xi)}$ and $p^{(\xi)}$.\\ 
Consider that the multi-object density at time $k$ is $\delta$-GLMB, so as we mentioned earlier, the propagated density in the prediction step at time $k+1$ is also $\delta$-GLMB, and is given as below:\\
\begin{equation}
	\label{eq:6}
	\boldsymbol{\pi}_{+}(\textbf{X})=\Delta(\textbf{X})
	\sum_{\substack{\xi \in \Xi}} \;\; \sum_{\substack{\{J,L_{+}\} \in \mathcal{F}(\mathbb{L}) \cup \mathcal{F}(\mathbb{B})}} \omega^{(J\cup L_{+},\xi)}_+\delta_{J \cup L_+}\big{[}\mathcal{L}(\textbf{X})\big{]} \Big{[}p^{(\xi)}_+\Big{]}^{\textbf{X}}
\end{equation}
where $\mathbb{B}$ is the discrete space of newly born objects' labels. Furthermore, $J \in \mathcal{F}(\mathbb{L})$ is the set of survived objects' labels and $L_+ \in \mathcal{F}(\mathbb{B})$ is the set of birthed objects' labels. The predicted weights and probability densities are given as below:
\begin{equation}
	\label{eq:7}
	\omega^{(J\cup L_{+},\xi)}_+ = 1_{\mathcal{F}(\mathbb{B})}(L_+) \; r^{L_+}_{B} \; \Big{[} 1-r_{B} \Big{]}^{\mathbb{B} - {L}_+} \times \sum_{\substack{I \in \mathcal{F}(\mathbb{L})}} 1_{\mathcal{F}(I)}(J) \; \Big{[} P^{(\xi)}_{S} \Big{]}^J \; \Big{[} 1 - P^{(\xi)}_{S} \Big{]}^{I-J} \omega^{(I,\xi)}
\end{equation}

\begin{equation}
	\label{eq:8}
	P^{(\xi)}_{S}(\ell)= \bigg{\langle} p^{(\xi)}(.,\ell) \: , \: P_S(.,\ell)\bigg{\rangle}
\end{equation}

\begin{equation}
	\label{eq:9}
	p^{(\xi)}_{+}(x_+,\ell_+) = 1_{\mathbb{L}}(\ell_+) \; \frac{\bigg{\langle} p^{(\xi)}(.,\ell_+) \: , \: P_S(.,\ell_+) f_+(x_+|.,\ell_+)\bigg{\rangle}}{P^{(\xi)}_{S}(\ell_+)} + 1_{\mathbb{B}}(\ell_+) \; p_{B}(x_+,\ell_+)
\end{equation}

where $P_S(x,\ell)$ is the probability that the object with label $\ell$ survives and $f_+(x_+|x,\ell_+)$ is the transition density from state $(x,\ell)$ to state $(x_+,\ell_+)$ while $\ell_+=\ell$. Moreover $r_{B}(\ell_+)$ is the probability of birth of a new object with label $\ell_+$ and $p_{B}(x_+,\ell_+)$ denotes its probability density on $\mathbb{X}$. Note that birth's RFS follows GLMB density in general and is given as below:\\
\begin{equation}
	\label{eq:10}
	f_B(\textbf{Y})=\Delta(\textbf{Y})\omega_B(\mathcal{L}(\textbf{Y})) \Big{[}p_B\Big{]}^\textbf{Y}
\end{equation}
Labeled multi-Bernoulli is an example of GLMB density with
\begin{equation}
	\label{eq:11}
	\omega_B(J) = \prod_{\substack{\{\ell\} \in \mathbb{B}}}(1-r_B(\ell)) \; \prod_{\substack{\{\ell'\} \in \mathbb{B}-J}}\frac{r_B(\ell')}{1-r_B(\ell')}
\end{equation}
where $J \in \mathbb{B}$. Note that in the implementation of $\delta$-GLMB filter, the birth's RFS follows Labeled multi-Bernoulli. \\
The posterior multi-object density at time $k+1$ has also the mathematical form of $\delta$-GLMB density under the condition of the standard measurement model. Note that the new discrete space of the objects' labels is $\mathbb{L_+}=\mathbb{L} \cup \mathbb{B} $. The updated multi-object density is given as below:
\begin{equation}
	\label{eq:12}
	\boldsymbol{\pi}_{Z}(\textbf{X}) \propto \Delta(\textbf{X})
	\sum_{\substack{(\xi,\theta) \in \Xi \cup \Theta}} \;\;\; \sum_{\substack{I_+ \in \mathcal{F}(\mathbb{L}_+)}} \omega^{(I_+,(\xi,\theta))}_{Z}\delta_{I_+}\big{[}\mathcal{L}(\textbf{X})\big{]} \Big{[}p^{(\xi,\theta)}_{Z}\Big{]}^{\textbf{X}}
\end{equation}
where $Z=\{z_1,z_2,...,z_{|Z|}\}$ is the set of measurements at time $k+1$ and $\Theta$ is the track to measurement association map at the current time. The weight and the probability density of the posterior $\delta$-GLMB density is given as follows:\\
\begin{equation}
	\label{eq:13}
	\omega^{(I_+,(\xi,\theta))}_{Z} = 1_{\Theta(I_+)}(\theta) \Big{[} \psi^{(\xi,\theta)}_{Z}\Big{]}^{I_+} \; \omega^{(I_+,\xi)}_+
\end{equation}

\begin{equation}
	\label{eq:14}
	\psi^{(\xi,\theta)}_{Z}(\ell) = \bigg{\langle} p^{(\xi)}_+ (.,\ell) \: , \: \psi^{(\theta(\ell)	)}_{Z}(.,\ell)\bigg{\rangle}
\end{equation}

\begin{equation}
	\label{eq:15}
	\psi^{(j)}_{Z}(x,\ell)=
	\begin{cases}
		\frac{P_D(x,\ell)g(z_j|x,\ell)}{\kappa(z_j)}, &  \; j \in \{1,...,|Z|\}\\
		1-P_D(x,\ell), &  \; j=0\\
	\end{cases}
\end{equation}

\begin{equation}
	\label{eq:16}
	p^{(\xi,\theta)}_{Z}(x,\ell)=\frac{p^{(\xi)}_+ (x,\ell) \: \psi^{(\theta(\ell)	)}_{Z}(x,\ell)}{\psi^{(\xi,\theta)}_{Z}(\ell)}
\end{equation}
where $P_D(x,\ell)$ denotes the probability of detection of the object $(x,\ell)$ and $g(z|x,\ell)$, and $\kappa(z)$ are respectively the likelihood and clutter intensity functions.
In this paper, as we explain in the following sections, we expand $\delta$-GLMB filter and utilize it in a new proposed model to effectively track multiple objects. Note that the implementation method we apply is introduced in \cite{vo2013labeled}. We use Gaussian-mixture implementation for propagating multiple-object density through the time. In order to do this Vo et al. applied K-shortest path and ranked assignment problem solvers respectively in prediction and update steps. As hinted in \cite{vo2016efficient}, we put limitation on the number of acceptable hypotheses and also define a lower band for the weight of acceptable hypotheses. Given a posterior density function, we can apply several methods to estimate the states of the objects. The joint Multi-Object Estimator and the Marginal Multi-Object Estimator are Bayes optimal choices that are not tractable. Alternatives are (maximum \emph{a posteriori}) MAP and (expected \emph{a posteriori})\ EAP methods. \\

\subsection{Proposed Birth Model}
\label{sec:3_2}

In this section we introduce our proposed birth model. As we mentioned in \ref{sec:3_1}, the birth RFS generally follows the GLMB density (\ref{eq:4}) and in this paper, we use labeled multi-Bernoulli density to implement $\delta$-GLMB filter (\ref{eq:5}). In literature, considering fixed points of the scene to be the places where new objects enter, is frequently chosen as the birth model. As a matter of fact, these points are the local maxima of the $p_B(x,\ell)$. Each specific point assigns a label to the new detected object. It worth to mention that each label, $(\ell_t,\ell_b)$, includes the stamps of time $(\ell_t)$ and place $(\ell_b)$ to make objects distinguishable. Thus, each measurement (as a birth candidate) will be weighted according to distance from the local maxima of the distribution. This model causes some major problems. In fact using a fixed distribution through the time, which is applied to all measurements, generates many false positives in prediction and update steps. In addition to the extra computational load induced, although most of these false positives are weak candidates, some of them possibly can cause disruption to track continuity. Also, predefined birth points (local maxima) may lead to miss birth tracks in successive frames if they are not sufficiently near the maxima. Additionally, predefined number of maxima restricts the number of simultaneous detected birth tracks. \\
In this paper we introduce a new birth model to address the aforementioned problems. To do this, we omit the restrictions (predefined and constant number of maxima) while the birth distribution generally follows the equation in (\ref{eq:10}). Note that we define the unassigned measurements of the preceding time step as the maxima of the distribution in the current time step. In fact unassigned measurements are candidates for birth points as they do not belong to the existing tracks. In order to apply this model, it is essential to wait a time step to select a preceding unassigned measurement as a birth point. So each newly born object will be recognized with a time step delay.  \\
At the cost of accepting a time step delay, reducing the computational load and applying more flexible birth model are the advantages we obtain. In fact, clutter removal is automatically performed by applying this model, since it is hardly possible that clutters continue to appear in a specific area in successive time steps. We also propose to apply a gating step in new measurement-to-birth point association part in order to weaken incorrect birth candidates. The proposed birth model is explained in Algorithm 1.\\

\begin{algorithm}
	
	\caption{Pseudocode for our proposed birth model.}
	\label{alg:1}
	\hspace*{\algorithmicindent} \textbf{\underline{\textsc{Inputs:}}} \\
	\hspace*{\algorithmicindent}   Unassigned measurements at time $k-1$: $Z_{unassigned,k-1}$ \\ 
	\hspace*{\algorithmicindent}   Measurements at time $k$: $Z_k$ \\ 
	\hspace*{\algorithmicindent}   Overlapping Threshold: $T_{overlapping}$ \\
	\hspace*{\algorithmicindent} \textbf{\underline{\textsc{Outputs:}}} \\
	\hspace*{\algorithmicindent}  Birth points at time $k$: $birth\_points$ \\ 
	\hspace*{\algorithmicindent}  Valid measurements at time $k$ for each birth point: $valid\_Z_k$ \\ 
	\begin{algorithmic}[1]
		\State{$birth\_points = \emptyset$}
		\For{$z \in Z_{unassigned,k-1}$}
		\State{$valid\_Z_k = \emptyset$}
		\For{$z' \in Z_k$}
		\If{$\textsc{IntersectionOverUnion}(z,z') > T_{overlapping}$}
		\State{$birth\_points = birth\_points \cup \{z\}$}
		\State{$valid\_Z_k = valid\_Z_k \cup \{z'\}$}			
		\EndIf
		\EndFor
		\State{$birth\_points = \textsc{Unique}(birth\_points)$}
		\Comment {Here, we save $valid\_Z_k$.}
		\EndFor
	\end{algorithmic}
\end{algorithm}

\begin{figure*}
	\centering
	
	\includegraphics[width=1\linewidth]{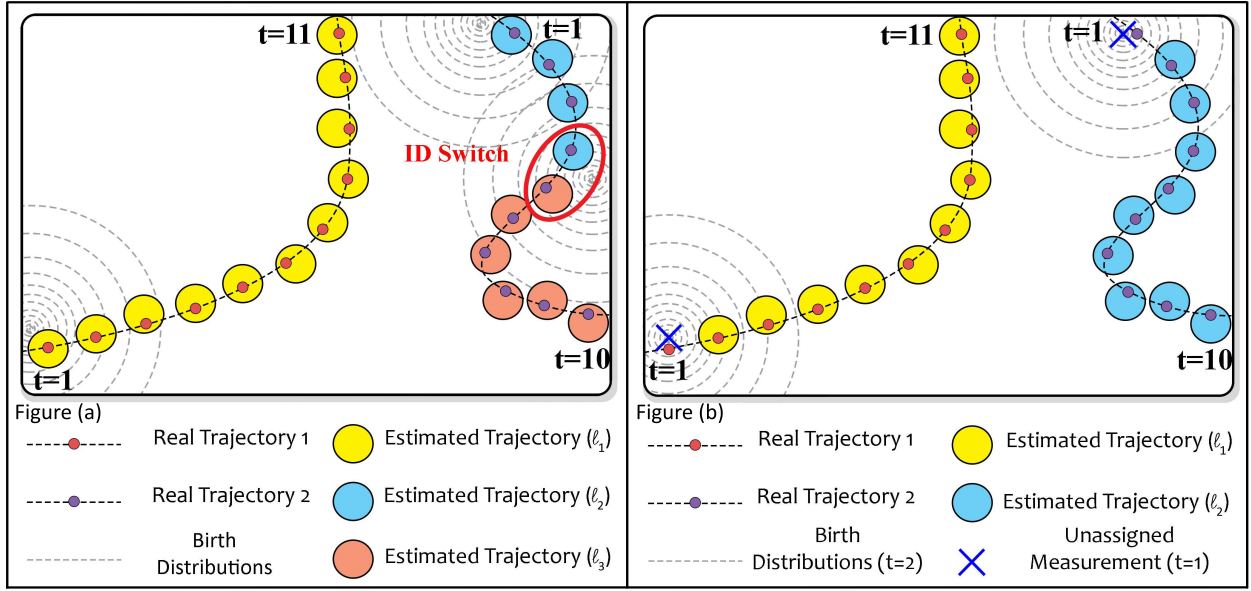}

	\caption{Illustration of the proposed birth model. (a) Predefined birth points are specified to detect newly-birthed objects. The red closed curve shows an ID switch occurred to object 2 because of approaching a constant and predefined birth point. (b) The birth model is applied and there is no constant birth point to distract ID consistency, so object 2 does not face ID switch.}
	\label{fig:2}
\end{figure*}

In fact each unassigned measurement at time $k-1$ $(z \in Z_{Unassigned,k-1})$ is defined as a birth point and subsequently as a predicted new track, if there is at least one current measurement $(z' \in Z_k)$ which substantially overlaps it. Note that in order to calculate the overlapping area, we utilize IOU metric. A user-defined threshold $(T_{overlapping})$ is applied to evaluate whether or not the condition is satisfied (see line 5). Note that for each birth point, there is a subset of $Z_k$, named valid measurements $(valid\_Z_k)$, which satisfies the condition. In the update step, we propose to use $(valid\_Z_k)$, not all current measurements $(Z_k)$, in order to perform measurement-to-track association for each predicted new track (birth point). This approach helps to reduce the computational load and weaken incorrect birth candidates. \\
Figure \ref{fig:2} illustrates that how using the general birth model leads to disruption in the tracking process. Obviously there is no problem when an existing object (Object 1) does not cross a predefined birth point. But it differs when it (Object 2) approaches one. In this case, strong object candidates can be generated in several sequent frames which lead to disruption in the tracking process. For instance, in Figure \ref{fig:2}.a, a new label is assigned to an existing object (Object 2). On the other hand, when we apply our proposed birth model, whether or not an object approaches a birth place, birth process is performed well. As it is shown in Figure \ref{fig:2}.b, birth distributions are not constant in number and location. Only a pair of sufficiently near measurements in two successive frames (the older one is unassigned), can generate a birth distribution. So the old measurement is defined as a strong temporary birth point.\\ 

\subsection{ID Switch Handler}
\label{sec:3_3}
As it was mentioned earlier, ID switch is a main problem in visual MOT applications. It occurs, when a pre-labeled object is estimated as a new trajectory. Although it is more probable that ID switch occurs after reappearance of occluded and miss-detected objects, it may happen for existing objects in the crowded scenes or due to the drawback of the state estimation section of the method. Note that, this is the incorrect chosen hypothesis returned by the state estimation section that leads to ID switch. In the following, we express typical implementation of state estimation section in $\delta$-GLMB filter. MAP estimator is performed on the cardinality distribution to obtain $\hat{N}$ (the estimated number of objects). Cardinality distribution is expressed as follows: \\
\begin{equation}
	\label{eq:17}
	\rho (n) = \sum_{\substack{(\xi,\theta) \in \Xi \cup \Theta}} \; \sum_{\substack{I_+ \in \mathcal{F}(\mathbb{L}_+)}} \omega^{(I_+,(\xi,\theta))}_{Z}\delta_{n}\big{[}|I_+|\big{]}
\end{equation}
and
\begin{equation}
	\label{eq:18}
	\hat{N} = \; \stackrel[n]{}{\mathrm{arg \; max}} \;\; \rho(n)
\end{equation}
Now, among hypotheses which include $\hat{N}$ objects, hypothesis with the highest weight is chosen:
\begin{equation}
	\label{eq:19}
	(\hat{I}_+,(\hat{\xi},\hat{\theta})) = 
	\stackrel[\tiny{\begin{matrix}
			I_+ \in \mathcal{F}(\mathbb{L}_+) \\ 
			(\xi,\theta) \in \Xi \cup \Theta
	\end{matrix}}]
	{}{\mathrm{arg \; max}} \;\; \omega^{(I_+,(\xi,\theta))}_{Z} \delta_{\hat{N}}\big{[}|I_+|\big{]} 
\end{equation}
So the estimated objects are defined as follows:
\begin{equation}
	\label{eq:20}
	\hat{\textbf{X}} = (\hat{x},\hat{\ell}) : \hat{\ell} \in \hat{I}_+ \; , \; \hat{x} = \int x \; p^{(\hat{\xi},\hat{\theta})}_{Z}(x,\hat{\ell}) \; dx
\end{equation}

In this paper we propose to continue using this estimator with new considerations. Algorithm 2 illustrates the pseudo code of the proposed method. After defining $\hat{N}$ and $\hat{\textbf{X}}$, we apply the size and color similarity functions ($\textsc{SameSize}$ and $\textsc{SameColor}$) in order to verify whether all of the estimated objects are correctly labeled or not. It is performed by using comparative user-defined thresholds $T_{size}$ and $T_{color}$.\\
If incorrect labeling is detected then we remove the chosen hypothesis from the pool of hypotheses and step back to $N$ estimation step. Note that elimination of a hypothesis changes the cardinality distribution and we also need to perform weight renormalization. So we continue repeating this process until all assigned labels are verified or the weight of the chosen hypothesis meets a user-defined threshold ($T_{\omega}$). In this case we keep the objects with verified labels as the estimated objects and the rest of objects will be categorized as miss-detected in the miss-detection handler section. \\
\emph{Remark}: Note that, in $\textsc{SameColor}$ function, we apply color histogram as similarity feature, so we can check similarity between two bounding boxes in two successive time steps based on histogram distance.\\

\begin{algorithm}
	
	\caption{Pseudocode for the ID switch handler and label recovery algorithm.}
	\label{alg:2}
	\hspace*{\algorithmicindent} \textbf{\underline{\textsc{Inputs:}}} \\
	\hspace*{\algorithmicindent}  Hypotheses at time $k$: $H = \{(I^{(h)},{\xi}^{(h)},{\omega}^{(h)},p^{(h)})\}^{N_h}_{h=1}$ \\ 
	\hspace*{\algorithmicindent}  Estimated hypothesis at time $k$: $\hat{h} = \{(I^{(\hat{h})},{\xi}^{(\hat{h})},{\omega}^{(\hat{h})},p^{(\hat{h})})\}$ \\ 
	\hspace*{\algorithmicindent}  Estimated states at time $k-1$: $\hat{\textbf{X}}_{k-1}=\{(x,\ell) : \ell \in \mathcal{L}(\hat{\textbf{X}}_{k-1}) \}$\\
	\hspace*{\algorithmicindent}  Estimated states at time $k$: $\hat{\textbf{X}}_{k}=\{(x,\ell) : \ell \in \mathcal{L}(\hat{\textbf{X}}_{k}) \}$\\
	\hspace*{\algorithmicindent}  Thresholds: $T_{size}, T_{color}, T_{\omega} $ \\
	\hspace*{\algorithmicindent}  \textbf{\underline{\textsc{Outputs:}}} \\
	\hspace*{\algorithmicindent}  Updated hypotheses at time $k$: $H^{U}=\{(I^{(h)},{\xi}^{(h)},{\omega}^{(h)},p^{(h)})\}^{N^{U}_h}_{h=1}$ \\ 
	\hspace*{\algorithmicindent}  Updated estimated hypothesis at time $k$: $\hat{h}^U = \{(I^{(\hat{h}^U)},{\xi}^{(\hat{h}^U)},{\omega}^{(\hat{h}^U)},p^{(\hat{h}^U)})\}$ \\ 
	\hspace*{\algorithmicindent}  Updated estimated states at time $k$: $\hat{\textbf{X}}^{U}_{k}=\{(x,\ell) : \ell \in \mathcal{L}(\hat{\textbf{X}}^{U}_{k}) \}$ \\ 
	
	\begin{algorithmic}[1]
		\State{$end{\_}while = 0$}
		\State{$\hat{h}^U =\hat{h}$} 
		\State{$H^U = H$} 
		\State{$\hat{\textbf{X}}^U_{k} = \hat{\textbf{X}}_{k}$}
		\While{$end{\_}while = 0$}
		\State{$end{\_}while = 1$}
		\For{$\ell \in \mathcal{L}(\hat{\textbf{X}}^U_{k})$}
		\If{$\ell \in \mathcal{L}(\hat{\textbf{X}}_{k-1}) \; \Andi \; end{\_}while = 1$} 
		\State{$\textbf{x}=(x,\ell) \; : \; ((x,\ell) \in \hat{\textbf{X}}^U_{k})$}
		\State{$\textbf{x}'=(x',\ell') \; : \; ((x',\ell') \in \hat{\textbf{X}}_{k-1})$}
		\If{$\textsc{SameSize}(x,x')\!<\!T_{size}\;  \Or \; \textsc{SameColor}(x,x')\!<\!T_{color}$}
		\State{$end{ \_}while = 0$}
		\Comment {Not similar in size or/and color.}
		\EndIf			
		\EndIf
		\EndFor
		\If{$end{\_}while = 0$}
		\State{$H^* = H^U - \hat{h}^U$} 
		\State{$\hat{N}=\textsc{MAPEstimator}(H^*)$}
		\State{$(\hat{h}^*,\hat{\textbf{X}^*})=\textsc{StateEstimator}(H^*,\hat{N})$}
		\If{$\omega_{\hat{h}}<T_{\omega}$}
		\State{$end{\_}while = 1$}
		\Else
		\State{$H^U =H^*$}
		\State{$\hat{h}^U = \hat{h}^*$}
		\State{$\hat{\textbf{X}}^{U}_{k} = \hat{\textbf{X}}^*$}
		\EndIf
		\EndIf
		\EndWhile
	\end{algorithmic}
\end{algorithm}

\subsection{Miss-Detection and Occlusion Handler}
\label{sec:3_4}
In this paper we propose a method to address occlusion and miss-detection. Applying separate specified GLMB filters for occluded and miss-detected objects (a GLMB filter for occlusion and a GLMB filter for miss-detection) is our solution to recover the disappeared tracks. Note that there are three main reasons why tracks disappear: Occlusion, miss-detection and leaving the scene. So disappeared tracks must be categorized into one of the three mentioned categories. Then a related task is performed accordingly. \\ 	

\begin{algorithm}
	
	\caption{Pseudocode for Categorizing disappeared Tracks.}
	\label{alg:3}
	\hspace*{\algorithmicindent} \textbf{\underline{\textsc{Inputs:}}}\\
	\hspace*{\algorithmicindent} Hypotheses at time $k$ returned from the ID Switch Handling module:\\ 
	\hspace*{\algorithmicindent} $H = \{(I^{(h)},{\xi}^{(h)},{\omega}^{(h)},p^{(h)})\}^{N_h}_{h=1}$ \\ 
	\hspace*{\algorithmicindent} Estimated states at time $k$ returned from the ID Switch Handling module:\\ 
	\hspace*{\algorithmicindent} $\hat{\textbf{X}}_{k}=\{(x,\ell) : \ell \in \mathcal{L}(\hat{\textbf{X}}_{k}) \}$\\
	\hspace*{\algorithmicindent} Estimated states at time $k-1$: $\hat{\textbf{X}}_{k-1}=\{(x,\ell) : \ell \in \mathcal{L}(\hat{\textbf{X}}_{k-1}) \}$\\
	\hspace*{\algorithmicindent} Lookup table of occluded objects: $LUT_{occ}$ \\ 
	\hspace*{\algorithmicindent} Lookup table of miss-detected objects: $LUT_{miss}$ \\ 
	\hspace*{\algorithmicindent} Overlapping Threshold: $T_{overlapping} $ \\
	\hspace*{\algorithmicindent} \textbf{\underline{\textsc{Outputs:}}} \\
	\hspace*{\algorithmicindent} Updated hypotheses at time $k$: $H^{U}=\{(I^{(h)},{\xi}^{(h)},{\omega}^{(h)},p^{(h)})\}^{N^{U}_h}_{h=1}$ \\ 
	\hspace*{\algorithmicindent} Updated lookup table of occluded objects: $LUT_{occ}$ \\ 
	\hspace*{\algorithmicindent} Updated lookup table of miss-detected objects: $LUT_{miss}$ \\ 
	\begin{algorithmic}[1]
		
		\For{$\ell \in \mathcal{L}(\hat{\textbf{X}}_{k-1})$}
		\If{$\ell \notin \mathcal{L}(\hat{\textbf{X}}_{k})$}
		\State{$\textbf{x}=(x,\ell) \; : \; ((x,\ell) \in \hat{\textbf{X}}_{k-1})$}
		\If{$\textsc{Location}(x) \in Border\_Zone \;\; \& \;\; \textsc{Direction}(x) = out$}
		\State{$H^{U}=\textsc{HypothesesRefinment}(H,\ell)$}
		\Else
		\State{$is\_occluded = 0$}
		\For{$\ell' \in \mathcal{L}(\hat{\textbf{X}}_{k-1}) \;\; \& \;\; \ell' \in \mathcal{L}(\hat{\textbf{X}}_{k})$}
		\If{$is\_occluded = 0$}
		\State{$\textbf{x}'=(x',\ell') \; : \; ((x',\ell') \in \hat{\textbf{X}}_{k})$}
		\State{$x_p=\textsc{Prediction}(x)$}
		\If{$\textsc{IntersectionOverArea}(x_p,x') > T_{overlapping}$}
		\State{$LUT_{occ}$ = $LUT_{occ} \cup \{(x,\ell,k)\}$}
		\State{$H^{U}=\textsc{HypothesesRefinment}(H,\ell)$}
		\State{$is\_occluded = 1$}
		\EndIf
		\EndIf
		\EndFor
		\If{$is\_occluded = 0$}
		\State{$LUT_{miss}$ = $LUT_{miss} \cup \{(x,\ell,k)\}$}
		\State{$H^{U}=\textsc{HypothesesRefinment}(H,\ell)$}
		\EndIf
		\EndIf
		\EndIf
		\EndFor	
		\State{}
		\Function {\textsc{HypothesesRefinment}} {$(H,\ell)$}
		\State{$H'=\emptyset$}
		\For{$h \in H$}
		\If{$\ell \in I(h)$}
		\State{$H' = H' \cup {h}$}
		\EndIf
		\EndFor
		\State{$H^{U} = H - H'$}
		\State{$H^{U} = \textsc{WeightNormalization}(H^U)$}
		\State\Return{$H^{U}$}
		\EndFunction
		
	\end{algorithmic}
\end{algorithm}

\begin{figure*}
	\centering
	
	\includegraphics[width=1\linewidth]{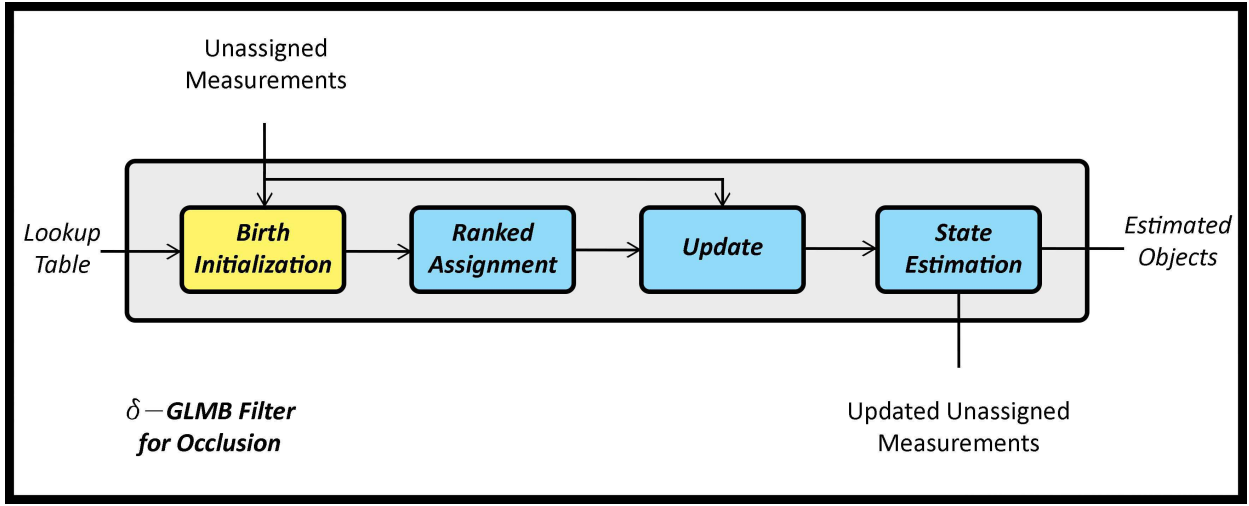}

	\caption{Illustration of $\delta$-GLMB for occlusion. A modified one-step $\delta$-GLMB filter is used to perform measurement-to-track association by considering all occluded objects in a frame.}
	\label{fig:3}
\end{figure*}

Algorithm 3 explains how we categorize disappeared objects. Firstly we check if any disappeared object has left the scene. Note that, when an object leaves the scene, it can confidently be said that it has been near the borders with direction into the out of the scene for the last few time steps. So we use \textsc{Location} and \textsc{Direction} functions in order to check this possibility. If a disappeared object does not satisfy this condition, then we need to figure out if it is an occluded or miss-detected object. To do this, we use the intersection over area (IOA) metric. IOA is formulated as follows:\\
\begin{equation}
	\label{eq:21}
	IOA(A,B) = \frac{area(A) \cap area(B)} {area(A)}
\end{equation}
Since we aim to calculate the overlapped area percentage for an object, we use IOA and not (for example) IOU. Note that, according to (\ref{eq:21}), by defining a threshold ($T_{overlapping}$), we can figure out whether the object $A$ is occluded by the object $B$ or not. Consider an object that is disappeared in the current time step. In order to define whether it is occluded or not, we check if it is overlapped with any currently existing object, while the disappeared object is presumed to be in its current predicted position. So we apply the \textsc{Prediction} function in which kinematic model is used to predict the position. Note that, if a disappeared object is not occluded and has not left the scene, we decide it is miss-detected.\\ 
In the situations of occlusion and miss-detection, it is possible for a disappeared object to reappear. So we need to make a lookup table containing its features in order to reuse it. Now, we can avoid a reappeared object to be recognized as a newly born one. Position, label and time of disappearance along with color properties are the main features we save. In Algorithm 3, we also apply \textsc{HypothesesRefinment} function that removes hypotheses containing disappeared objects. We do this as a part of our proposed method of occlusion and miss-detection handling.   \\ 
As it is shown in Figure \ref{fig:1}, after categorizing disappeared objects and creating lookup tables, we apply separate $\delta$-GLMB filters for occlusion and miss-detection. By this we aim to detect any reappeared object at time step $k$ and add it to the group of estimated objects returned by the main $\delta$-GLMB filter at this time. To clarify the procedure, we need to describe different parts of the filters. Due to the similarity, it suffices to describe one of them.\\
Firstly, we propose to use unassigned measurements (returned by the main $\delta$-GLMB filter) as the input measurements of the $\delta$-GLMB filter for occlusion. The reason is that the main $\delta$-GLMB filter is not designed to handle occlusion and miss-detection. Thanks to one time step delay in the process of detecting newly-birthed object in the main $\delta$-GLMB filter, we are allowed to claim that unassigned measurements are related to probable reappeared objects. \\
Secondly, we discuss about the applied birth model in $\delta$-GLMB filter for occlusion. In fact, $\delta$-GLMB filter for occlusion is not supposed to have a sequential nature and it is a one time-step filter. So we propose to use the general (not proposed) birth model mentioned is section \ref{sec:3_2}, since we do not have to accept a time step delay to detect a birth object. As a matter of fact, each estimated object (reappeared object) returned by the $\delta$-GLMB filter for occlusion, is a newly born object which takes the label of an occluded object. Suppose that $\mathbb{L}_{occ}$ is the space of occluded objects' labels which are currently present in the lookup table. Considering (\ref{eq:10}), we propose the following density as the birth density.   \\
\begin{equation}
	\label{eq:22}
	f_B(\textbf{Y})= \Delta(\textbf{Y})\sum_{\substack{I \in \mathbb{L}_{occ}}} \sum_{\substack{\upsilon \in \Upsilon}}\omega^{(I,\upsilon)}_B \; \delta_{I}\big{[}\mathcal{L}(\textbf{Y})\big{]} \; \Big{[}p^{(\upsilon)}_B\Big{]}^\textbf{Y}
\end{equation}
where $I$ is a subset of $\mathbb{L}_{occ}$, $\upsilon \in \Upsilon$ is a subset of unassigned measurement-to-birth point association map and $p^{(\upsilon)}_B$ is the kinematic state distribution of a birth point. In fact each unassigned measurement takes each label of occluded objects to create birth point candidates.\\
Additionally $\omega^{(I,\upsilon)}_B$ is a non-negative weight associated with label set $I$ and discrete index $\upsilon$, while $\sum_{\substack{I \in \mathbb{L}_{occ}}} \; \sum_{\substack{\upsilon \in \Upsilon}}\;\omega^{(I,\upsilon)}_B = 1$ and   \\
\begin{equation}
	\label{eq:23}
	\omega^{(I,\upsilon)}_B \propto 
	\prod_{\substack{\{\ell\} \in \mathbb{L}_{occ}} \atop \substack{\{\ell\} \notin I}}\omega_{\ell0}
	\prod_{\substack{\{\ell\} \in I}\atop \substack{m \in \upsilon(I)}}\omega_{\ell m}
\end{equation}
where $\upsilon(I) \in \{0,1,...,M\}$ and $M$ is the number of unassigned measurements. Moreover\\
\begin{equation}
	\label{eq:24}
	\sum_{m =0}^{M} \omega_{\ell m} = 1
\end{equation}
where
\begin{equation}
	\label{eq:25}
	\omega_{\ell 0} = 1 - r^{(\ell)} \; , \; \sum_{m =1}^{M} \omega_{\ell m} = r^{(\ell)}
\end{equation}
and $r^{(\ell)}$ is the probability of reappearing an occluded object with label $\ell$.\\
In order to define $\omega_{\ell m}$, we suggest to consider similarities between the occluded object (with label $\ell$) and the observed measurement (with index $m$). Angle and magnitude of velocity vector along with color properties are the features we use to measure the similarity. To do this, we propose a new distance metric in which we use direction-aware distance, introduced in \cite{gu2017new}. \cite{gu2017new} proposed a distance metric which is a combination of Euclidean distance and cosine similarity and includes both magnitude and angle of vectors. According to this metric, distance between vectors $\vec{x}$ and $\vec{y}$ is give as below:\\ 
\begin{equation}
	\label{eq:26}
	d_{DA}(\vec{x} , \vec{y})=\sqrt{\lambda_1^2\|\vec{x}-\vec{y}\|^2+\lambda_2^2(1-\frac{\vec{x}.\vec{y}}{\|\vec{x}\| \|\vec{y}\|})}
\end{equation}
where $\lambda_1$ and $\lambda_2$ are defined to adjust the importance allocated to each part of the metric. In fact we suggest to utilize the velocity vector similarity in order to perform directional search over the area in order to recover reappeared objects.   \\
We also use color as a feature of similarity. In visual MOT we expect that an absent object does not change drastically in its appearance in several frames. Accordingly we use color histogram so we can check similarity between two bounding boxes (before and after the occlusion or miss-detection) based on histogram distance. As mentioned in \ref{sec:3_3}, we use HSV and Bhattacharyya distance respectively as color space and distance metric. Consequently, $\omega_{\ell m}$ is expressed as:  \\
\begin{equation}
	\label{eq:27}
	\omega_{\ell m} \propto \exp(-\alpha(k-k_0)) \exp(-\frac{d^2_{DA}(\vec{v}_{k_0} , \vec{v}_{k_0:k})}{2\sigma_v^2}) + \exp(-\frac{d_B^2(H_{k}-H_{k_0})}{2\sigma_{H}^2})
\end{equation}
where $\vec{v}_{k_0}$ and $\vec{v}_{k_0:k}$ are respectively the vector velocity at time step $k_0$ (occlusion start time step) and the average velocity vector over the period $k_0$ to $k$. Also $H_{k_0}$ and $H_K$ denote HSV color histograms. $\sigma_{v}$ and $\sigma_{H}$ are user-defined parameters that scale the distances. Note that, in the first term, $\exp(-\alpha(k-k_0))$ is a forgetting factor which allows to decrease the importance of velocity similarity over the time since we are not confident that the object keeps the same velocity.   \\
In order to perform measurement-to-birth point association, we need to calculate $\omega^{(I,\upsilon)}_B$ (\ref{eq:23}). Exhaustive enumeration is needed to calculate the normalization factor, so we suggest to apply ranked assignment method (similar to update step implementation) to make the task tractable.\\
Figure \ref{fig:3} demonstrates the flowchart of the $\delta$-GLMB filter for occlusion in a modular form. Unassigned measurements returned from the previous part are used by both birth initialization and update section. Since the filter is a one-time step filter, there is no survived object and subsequently prediction concept. The rank assignment part accomplishes the task of creating strong hypotheses and the update step performs data association to update the hypotheses. Finally, the estimation section extracts the states of new objects (here occluded ones). The reason we suggest to use a GLMB filter-based module to recover the occluded objects is the advantage of considering multiple hypotheses in the situation of challenging occlusion scenarios such as adjacent occluded objects and finding the correct hypothesis based on an efficient framework. \\
\emph{Remark}: As we mentioned earlier, the $\delta$-GLMB filter for miss-detection works the same. As illustrated in Figure \ref{fig:1}, it is applied after the $\delta$-GLMB filter for occlusion and uses the unassigned measurements returned from this filter as input measurements. 

\section{Experimental Results}
\label{sec:4}
In this section, we first elaborate on the motion and measurement models in section \ref{sec:4_1}. Also the quantitative and qualitative experimental results taken by applying our method to well-known test datasets are presented in section \ref{sec:4_2}.
\subsection{Motion and Measurement Models, Parameter Setting}
\label{sec:4_1}
The most applied method of representing an object in MOT field is based on using bounding boxes. So for each object, here we can say person since we are focused on pedestrian tracking, a bounding box is supposed to define the minimized area which completely includes the object. It is a proper choice since most of well-known test datasets follow this standard. So along with position and velocity, box dimensions (width and height) are other parameters which are necessary to be included in state equation. So the labeled single-object state is formulated as $\textbf{X} = (x,\ell) = ([p_x,p_y,v_x,v_y,w,h],\ell)$ where $w$ and $h$ are respectively width and height of the corresponding bounding box. Note that we apply Gaussian mixture implementation of $\delta$-GLMB filter so it is required to consider standard linear Gaussian model for a single object. Each object follows a linear dynamic model and measurement model also is a linear Gaussian one as given below:
\begin{equation}
	\label{eq:28}
	f_{k|k-1}(x|x')=\mathcal{N}(x;F_{k-1}x',Q_{k-1})
\end{equation}
\begin{equation}
	\label{eq:29}
	g_k(z|x)=\mathcal{N}(z;H_{k}x,R_{k})
\end{equation}
where $F_{k-1}x'$ and $H_{k}x$ are the mean and $Q_{k-1}$ and $R_{k}$ are the covariance of the densities. $F_{k-1}$ and $H_k$ are given by:
\begin{eqnarray}
	\label{eq:30}
	F_{k-1}=
	\begin{bmatrix}
		1&0&\Delta t&0&0&0\\
		0&1&0&\Delta t&0&0\\
		0&0&1&0&0&0\\
		0&0&0&1&0&0\\
		0&0&0&0&1&0\\
		0&0&0&0&0&1\\
	\end{bmatrix}
	,\,\,\,\,\,H_k=
	\begin{bmatrix}
		1&0&0&0&0&0\\
		0&1&0&0&0&0\\
		0&0&0&0&1&0\\
		0&0&0&0&0&1\\
	\end{bmatrix}
\end{eqnarray}

\subsection{Qualitative and Quantitative Results}
\label{sec:4_2}

\begin{figure*}
	\centering
	\tiny
	\includegraphics[width=.5\linewidth]{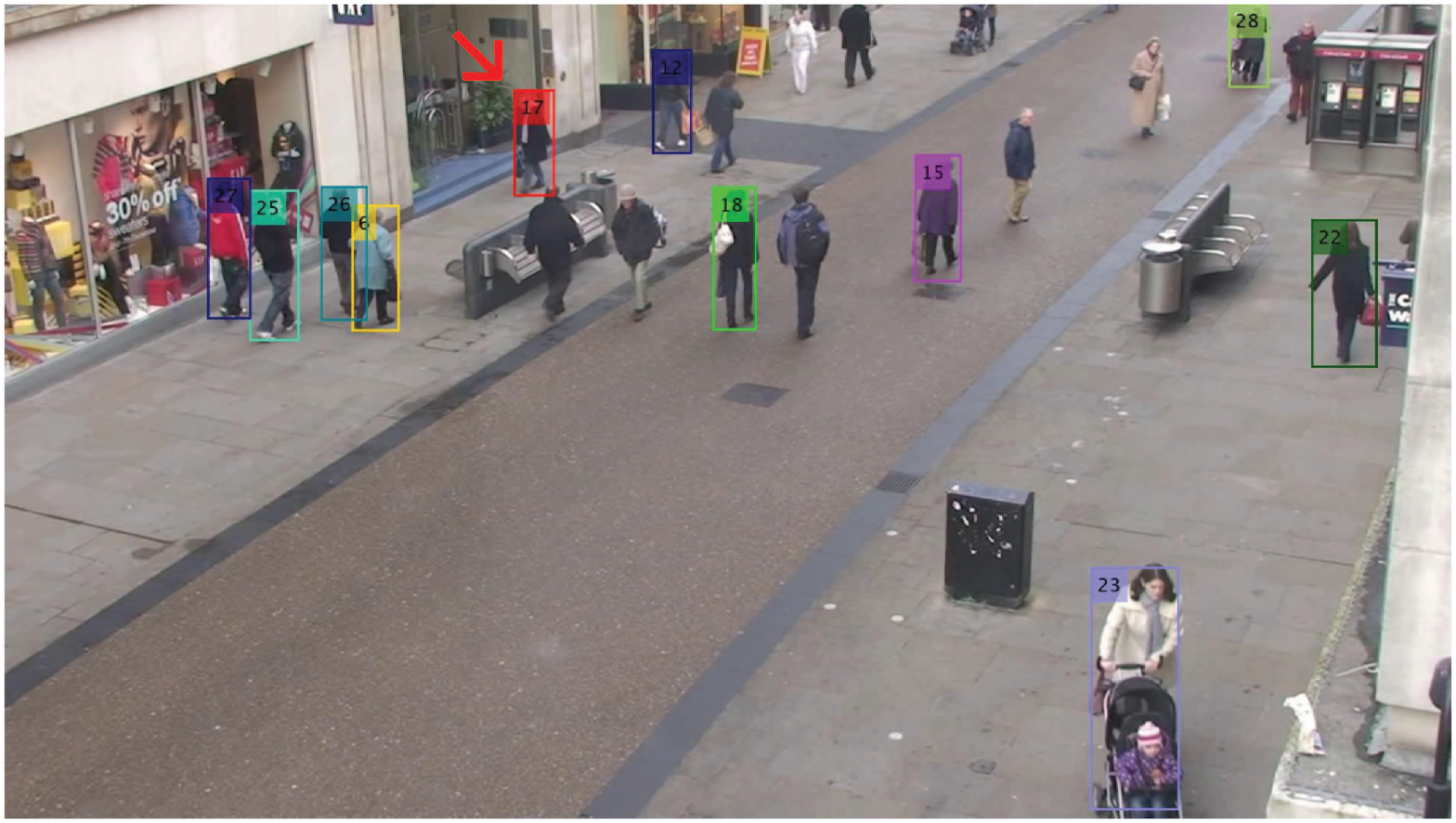}\hspace*{1em}
	\includegraphics[width=.5\linewidth]{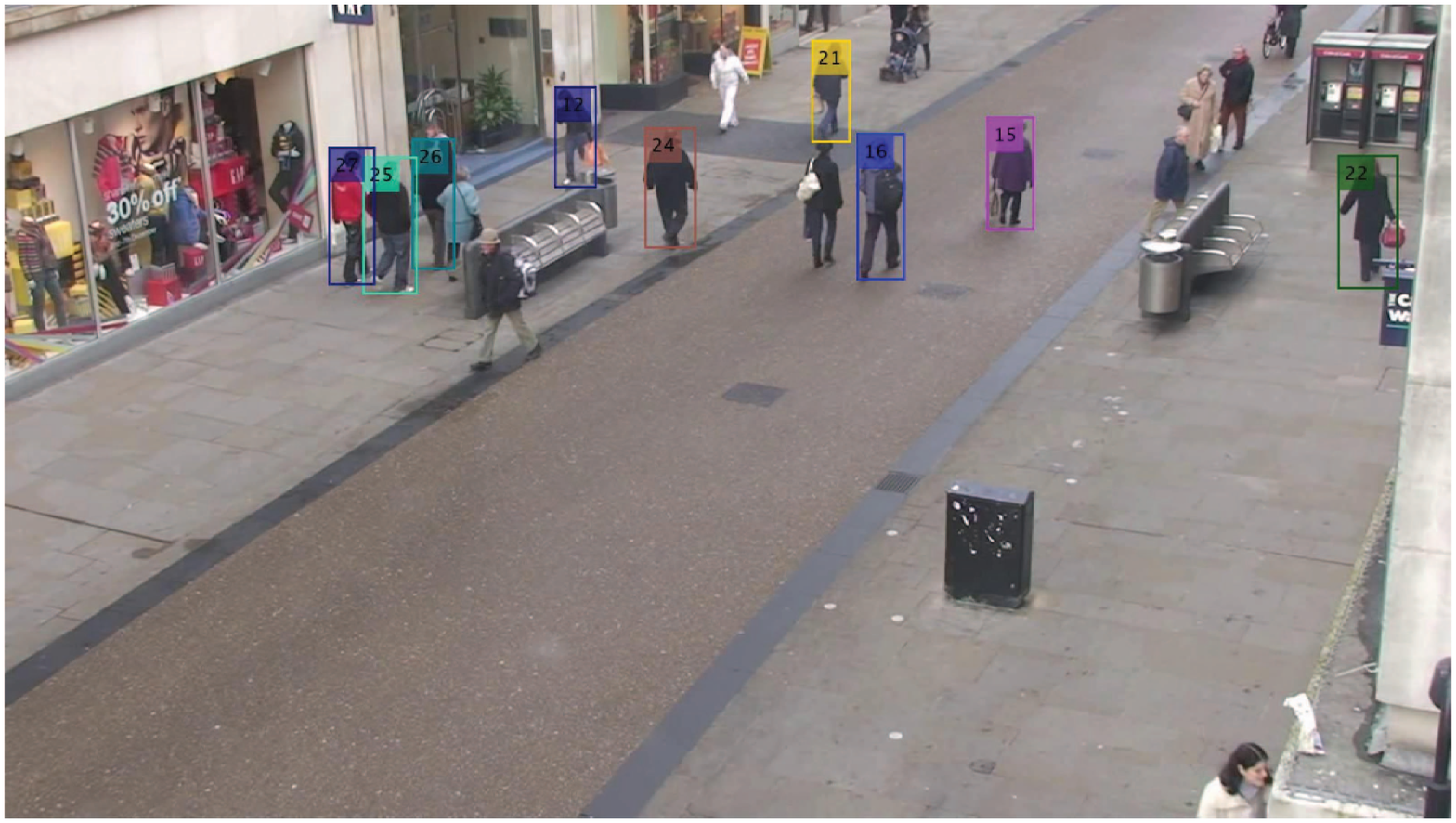}\\
	\vspace*{-2em}
	\textcolor{white}{\textbf{Frame 54}} \hspace*{35em}  \textcolor{white}{\textbf{Frame 59}}\\
	\vspace*{1em}
	\includegraphics[width=.5\linewidth]{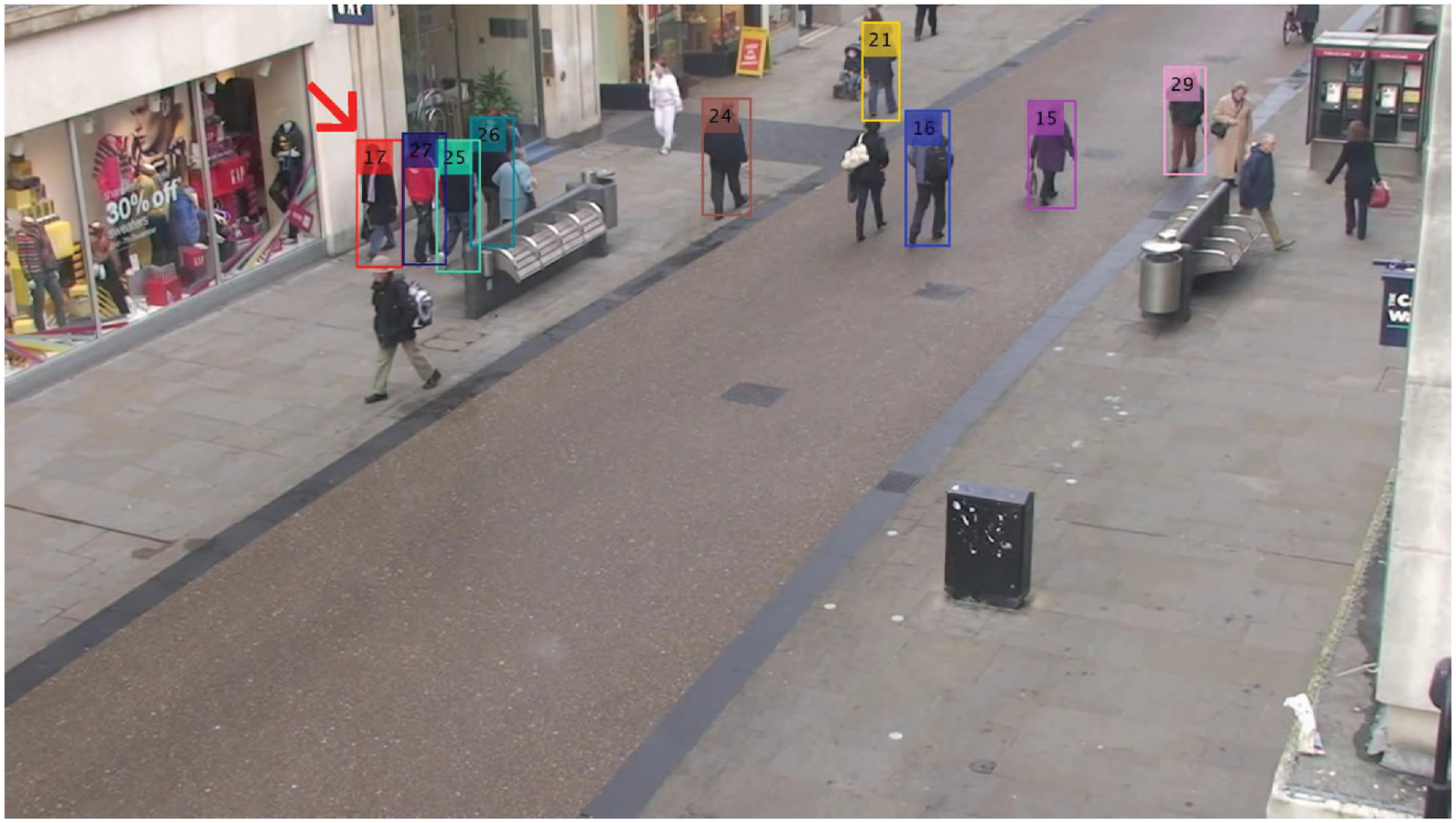}\hspace*{1em}
	\includegraphics[width=.5\linewidth]{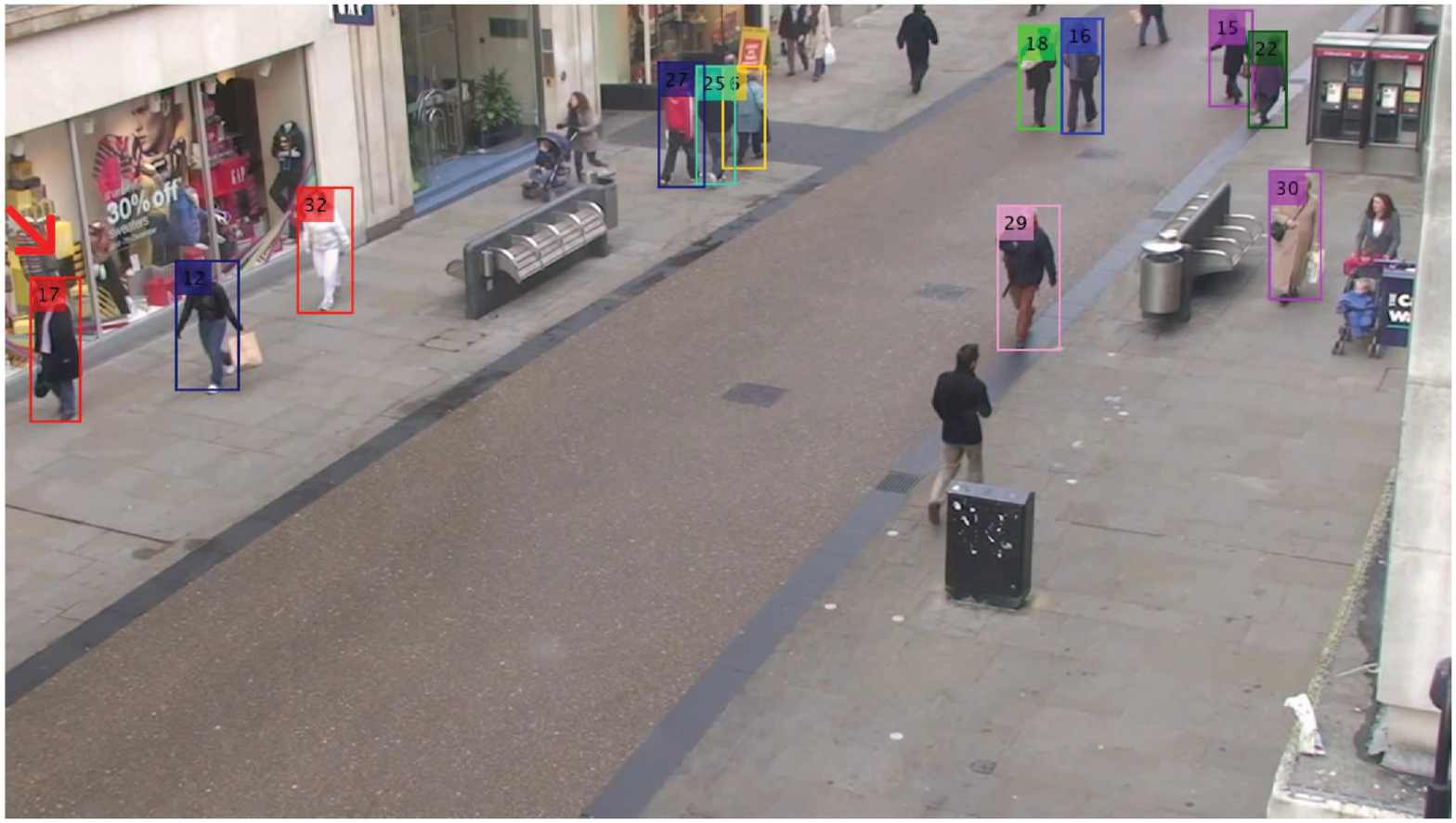}\\
	\vspace*{-2em}
	\textcolor{white}{\textbf{Frame 62}} \hspace*{35em}  \textcolor{white}{\textbf{Frame 76}} \\ 
	\vspace*{1em}
	(a)\\
	\vspace*{1em}
	\includegraphics[width=.5\linewidth]{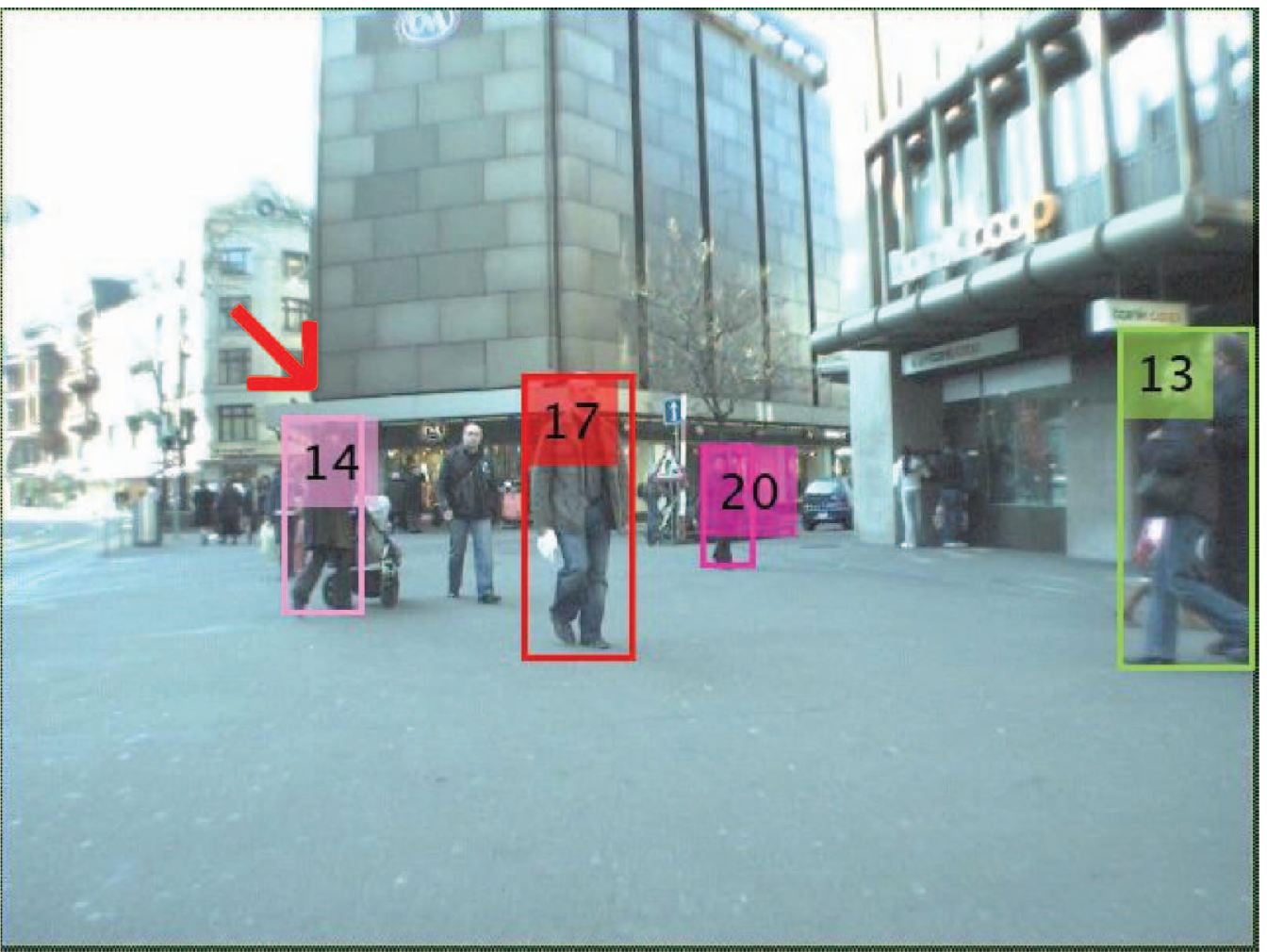}\hspace*{1em}
	\includegraphics[width=.5\linewidth]{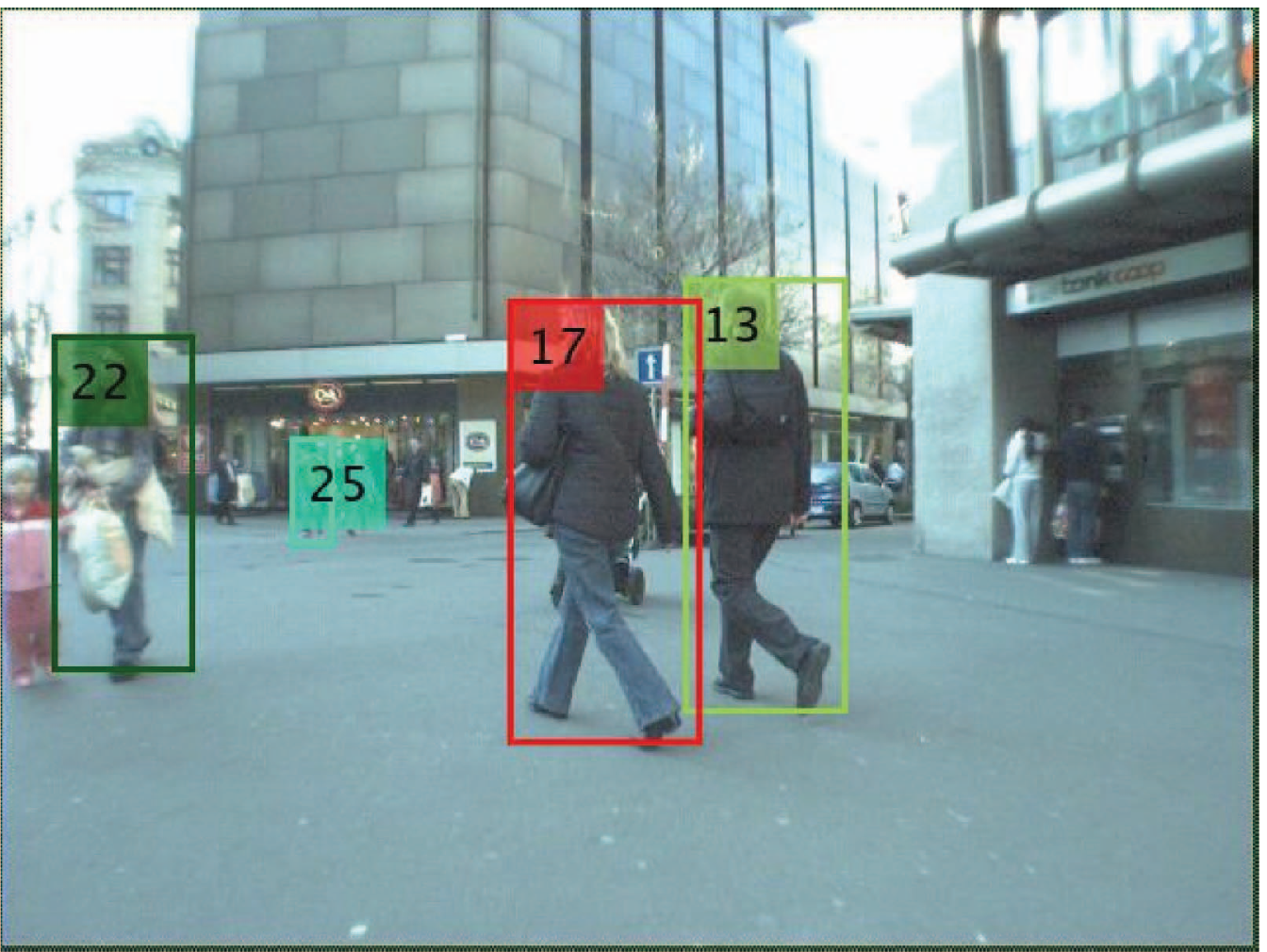}\\
	\vspace*{-2em}
	\textcolor{white}{\textbf{Frame 197}} \hspace*{35em}  \textcolor{white}{\textbf{Frame 261}}\\
	\vspace*{1em}
	\includegraphics[width=.5\linewidth]{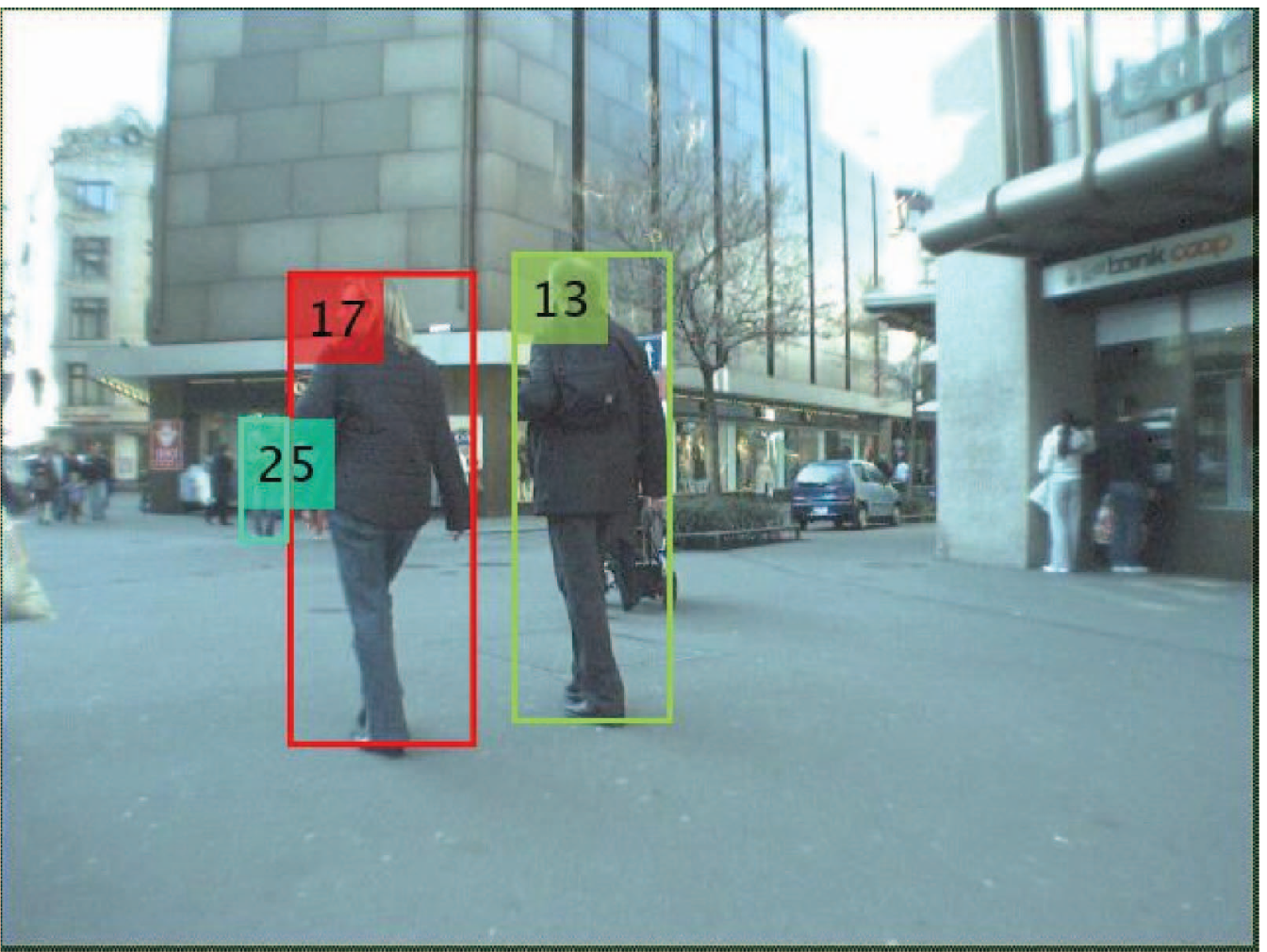}\hspace*{1em}
	\includegraphics[width=.5\linewidth]{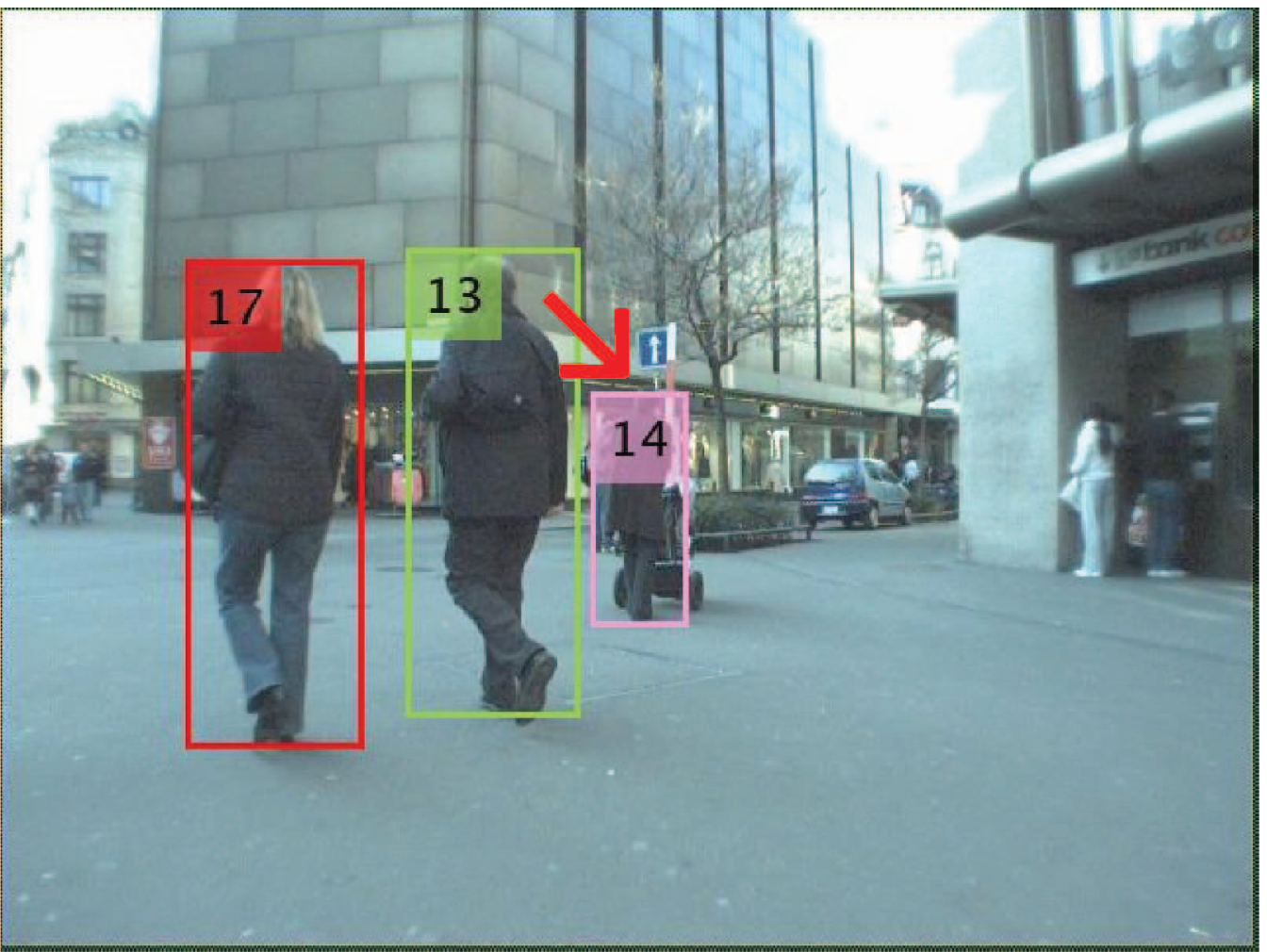}\\
	\vspace*{-2em}
	\textcolor{white}{\textbf{Frame 274}} \hspace*{35em}  \textcolor{white}{\textbf{Frame 281}} \\ 
	\vspace*{1em}
	(b)\\
	
	\caption{Illustration of qualitative results on MOT15 dataset. (a) Four sample frames of AVG-TownCentre sequence. (b) Four sample frames of ETH-Jelmoli sequence. In each part, (a) and (b), there is an occluded person that is correctly labeled after reappearance.}
	\label{fig:4}
\end{figure*}

\begin{figure*}
	\centering
	\tiny
	\includegraphics[width=.5\linewidth]{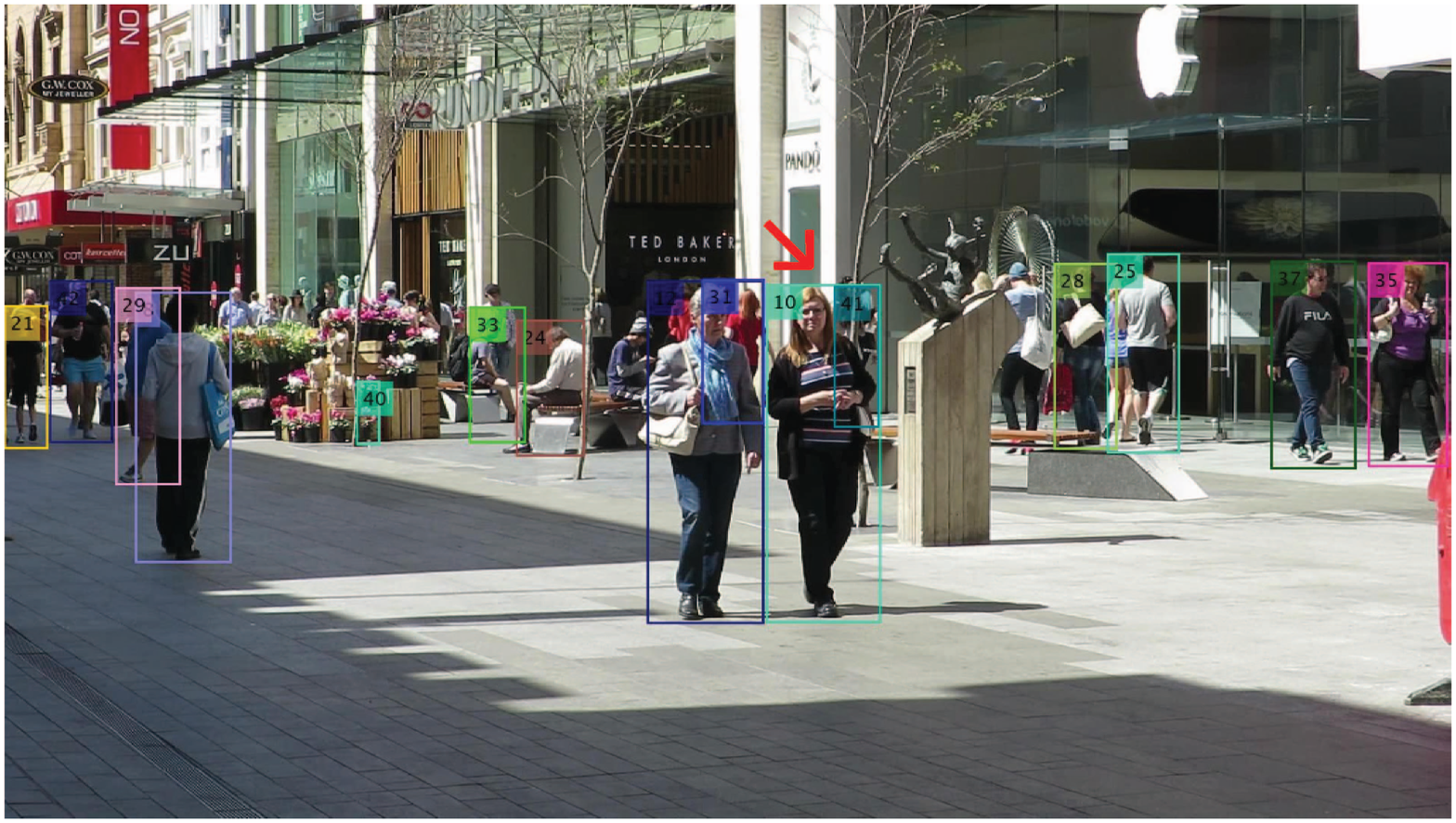}\hspace*{1em}
	\includegraphics[width=.5\linewidth]{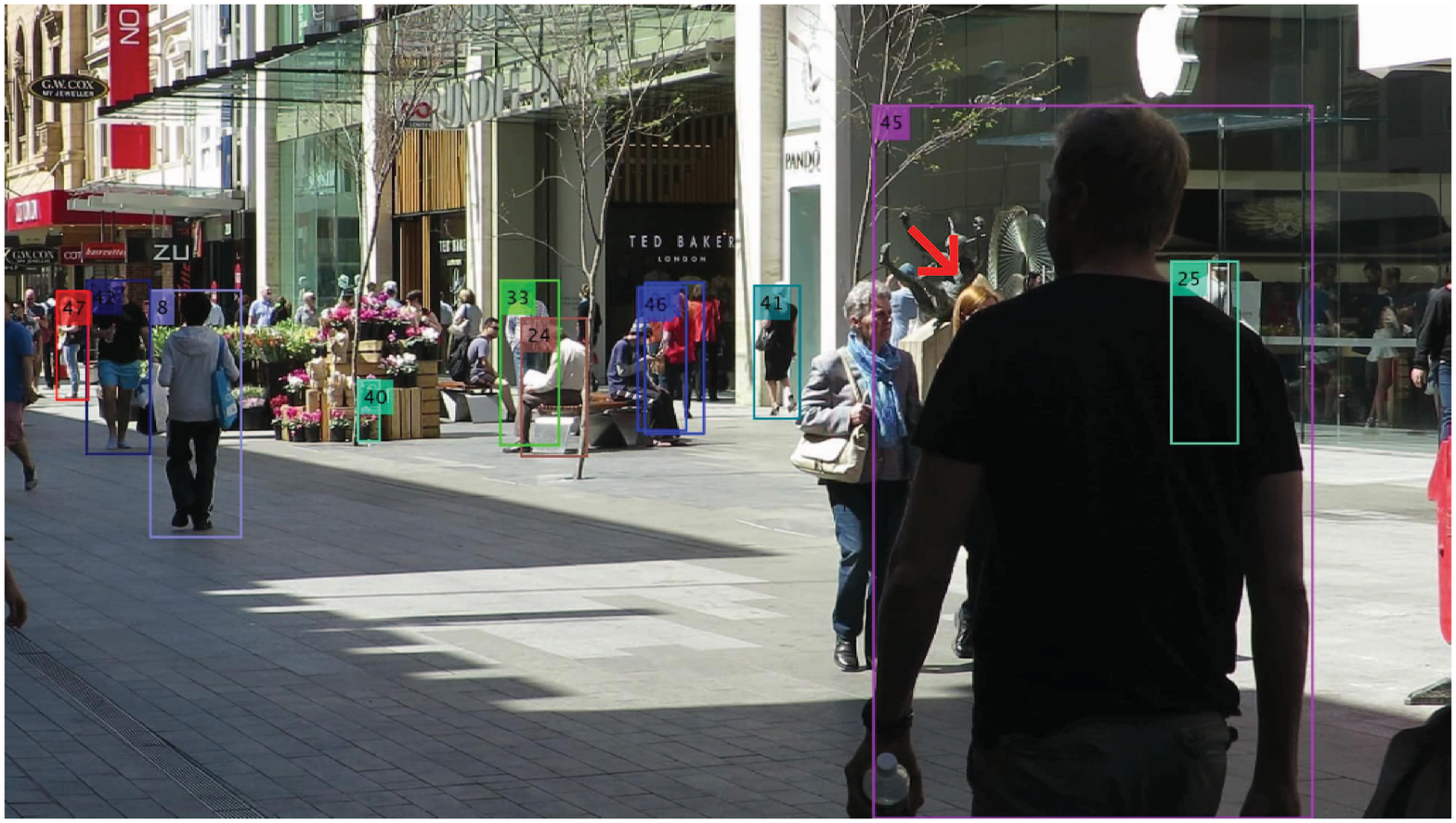}\\
	\vspace*{-2em}
	\textcolor{white}{\textbf{Frame 325}} \hspace*{35em}  \textcolor{white}{\textbf{Frame 383}}\\
	\vspace*{1em}
	\includegraphics[width=.5\linewidth]{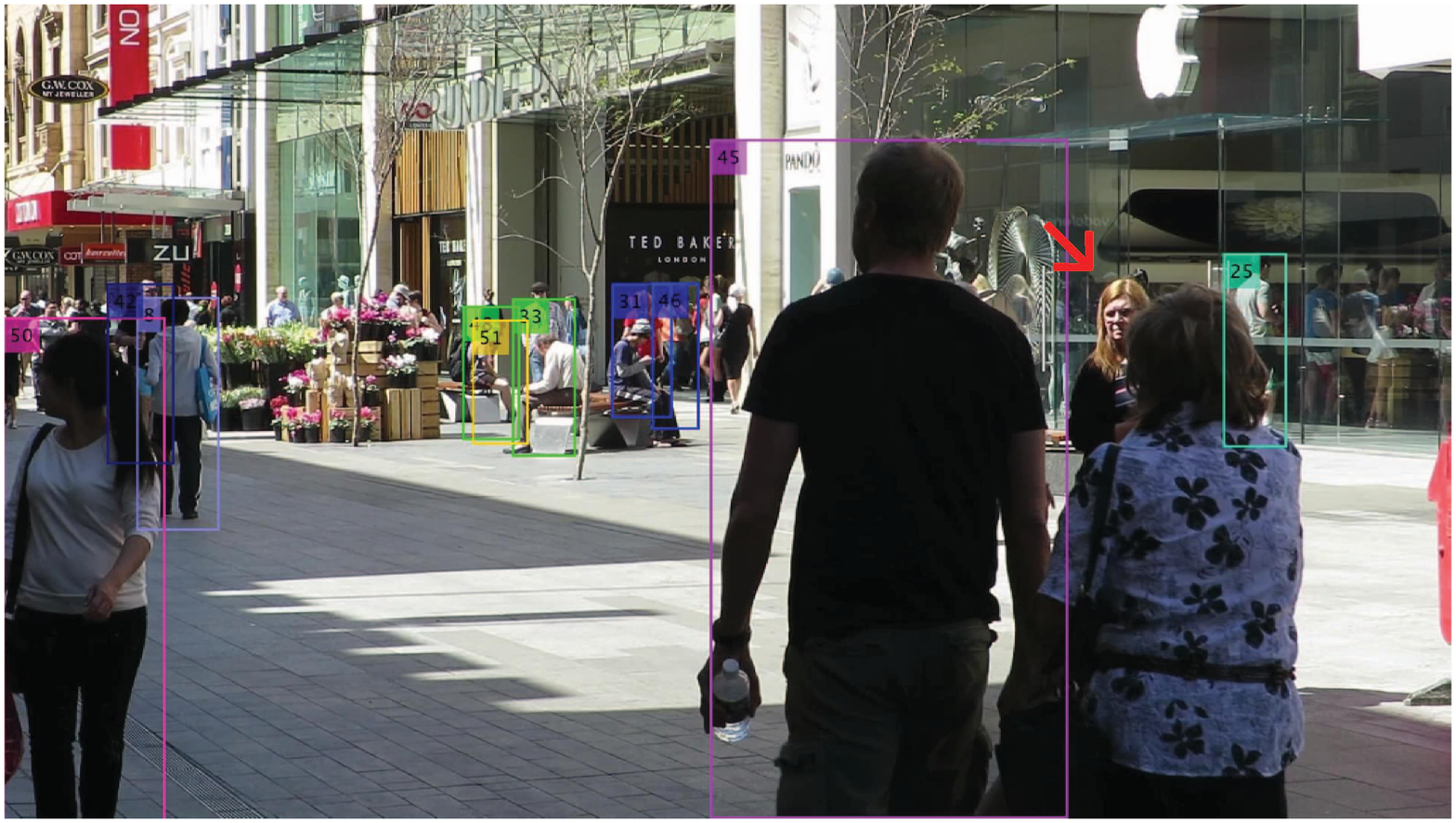}\hspace*{1em}
	\includegraphics[width=.5\linewidth]{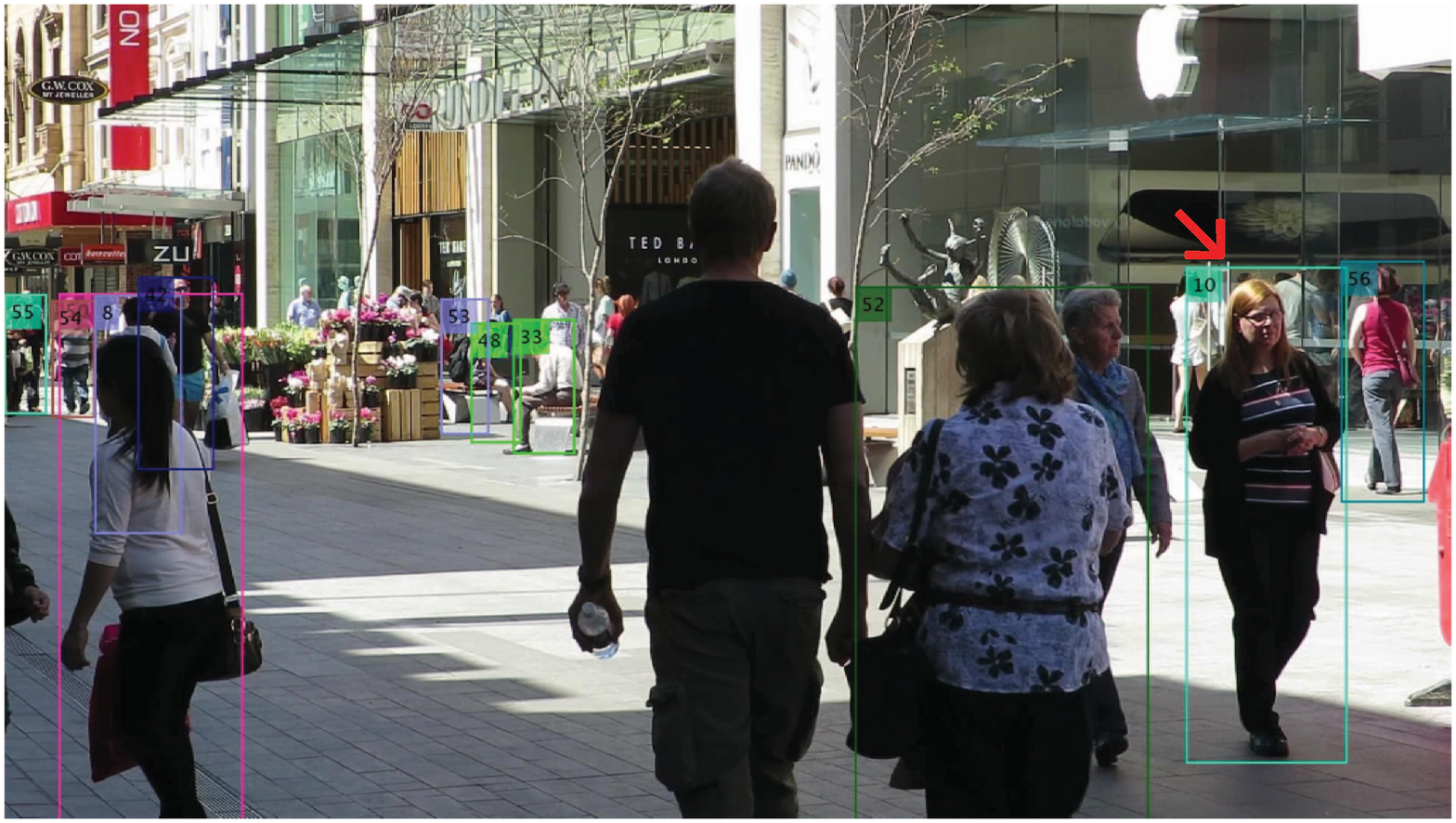}\\
	\vspace*{-2em}
	\textcolor{white}{\textbf{Frame 415}} \hspace*{35em}  \textcolor{white}{\textbf{Frame 442}} \\ 
	\vspace*{1em}
	(a)\\
	\vspace*{1em}
	\includegraphics[width=.5\linewidth]{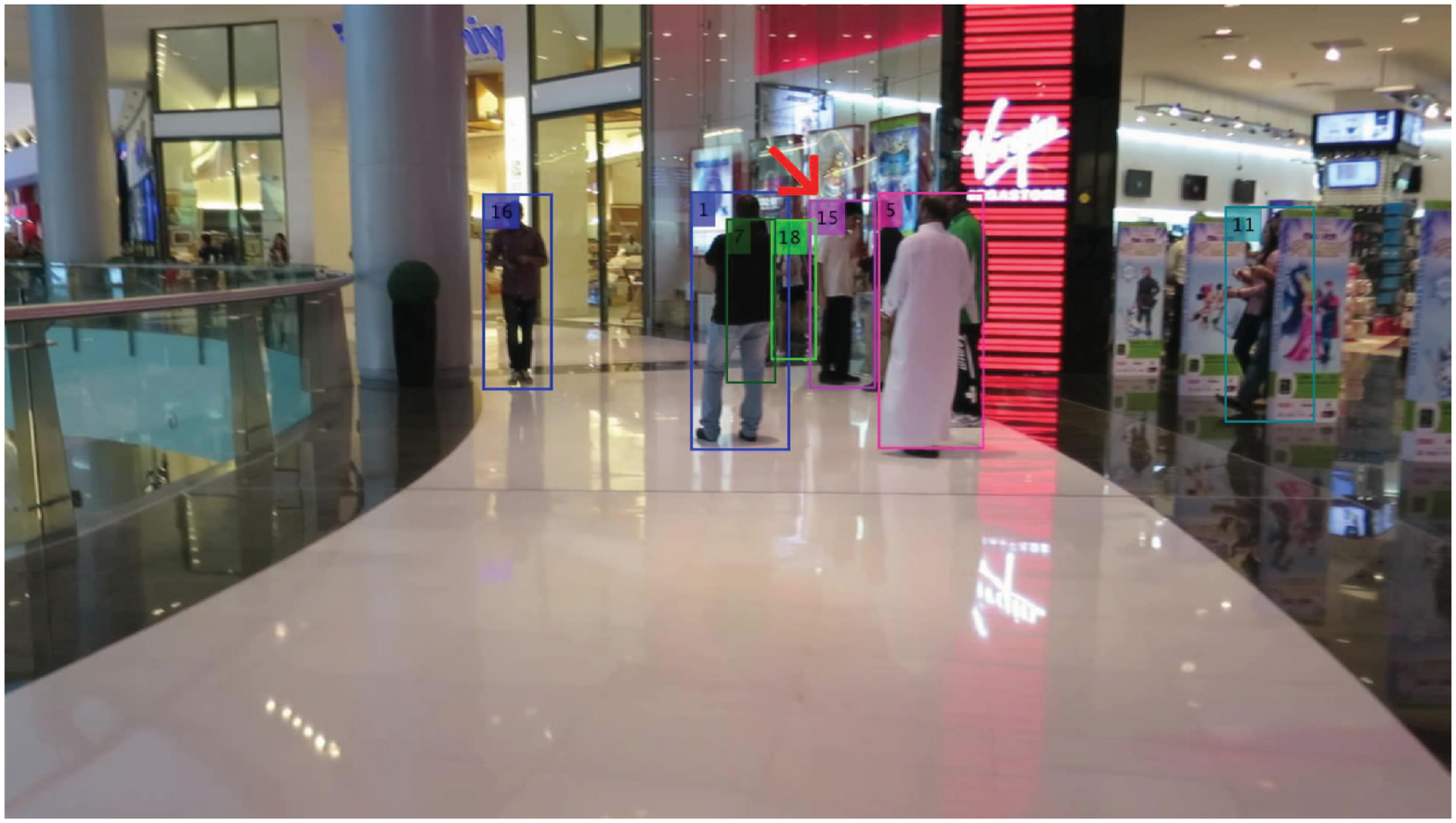}\hspace*{1em}
	\includegraphics[width=.5\linewidth]{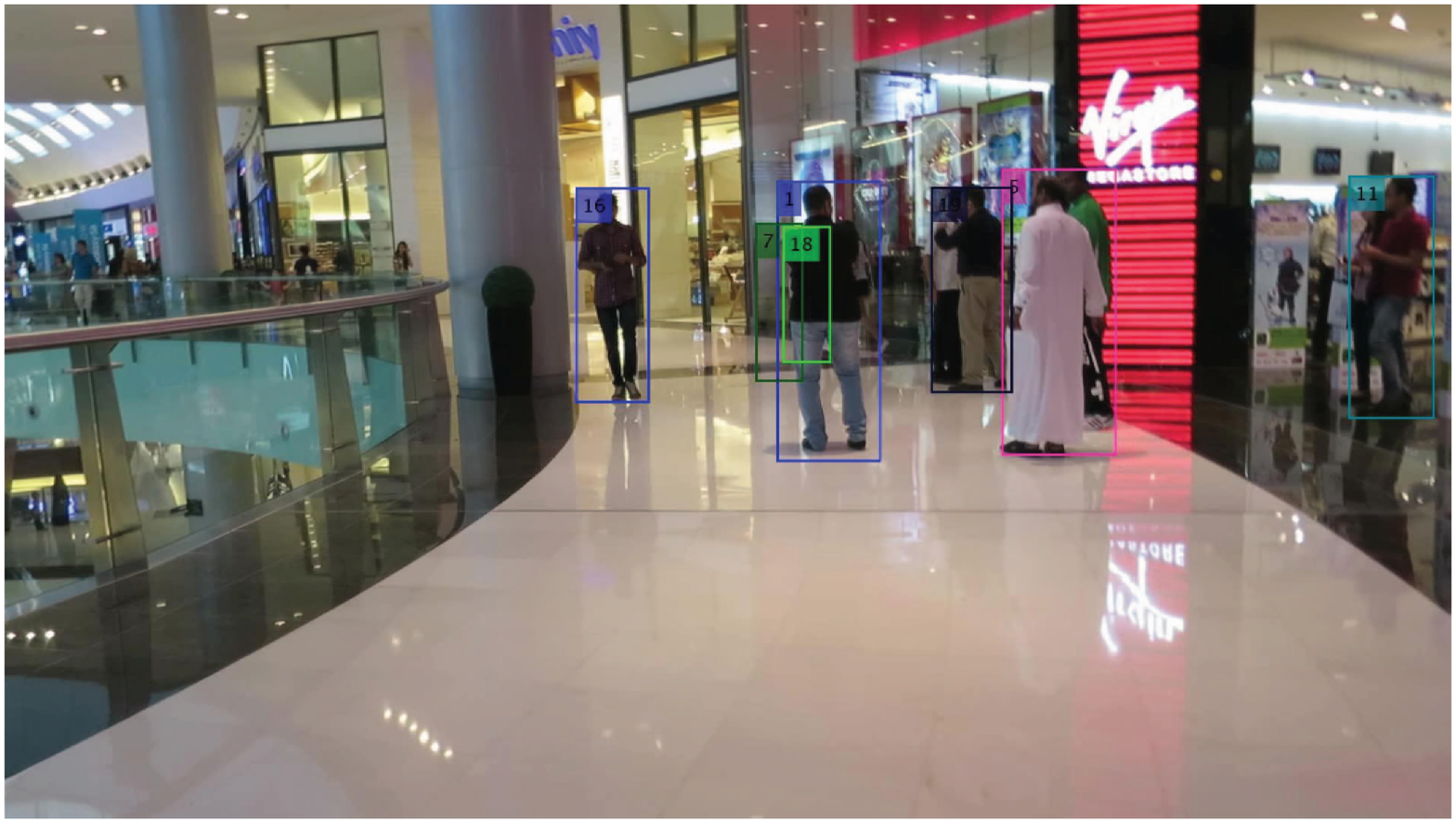}\\
	\vspace*{-2em}
	\textcolor{white}{\textbf{Frame 109}} \hspace*{35em}  \textcolor{white}{\textbf{Frame 123}}\\
	\vspace*{1em}
	\includegraphics[width=.5\linewidth]{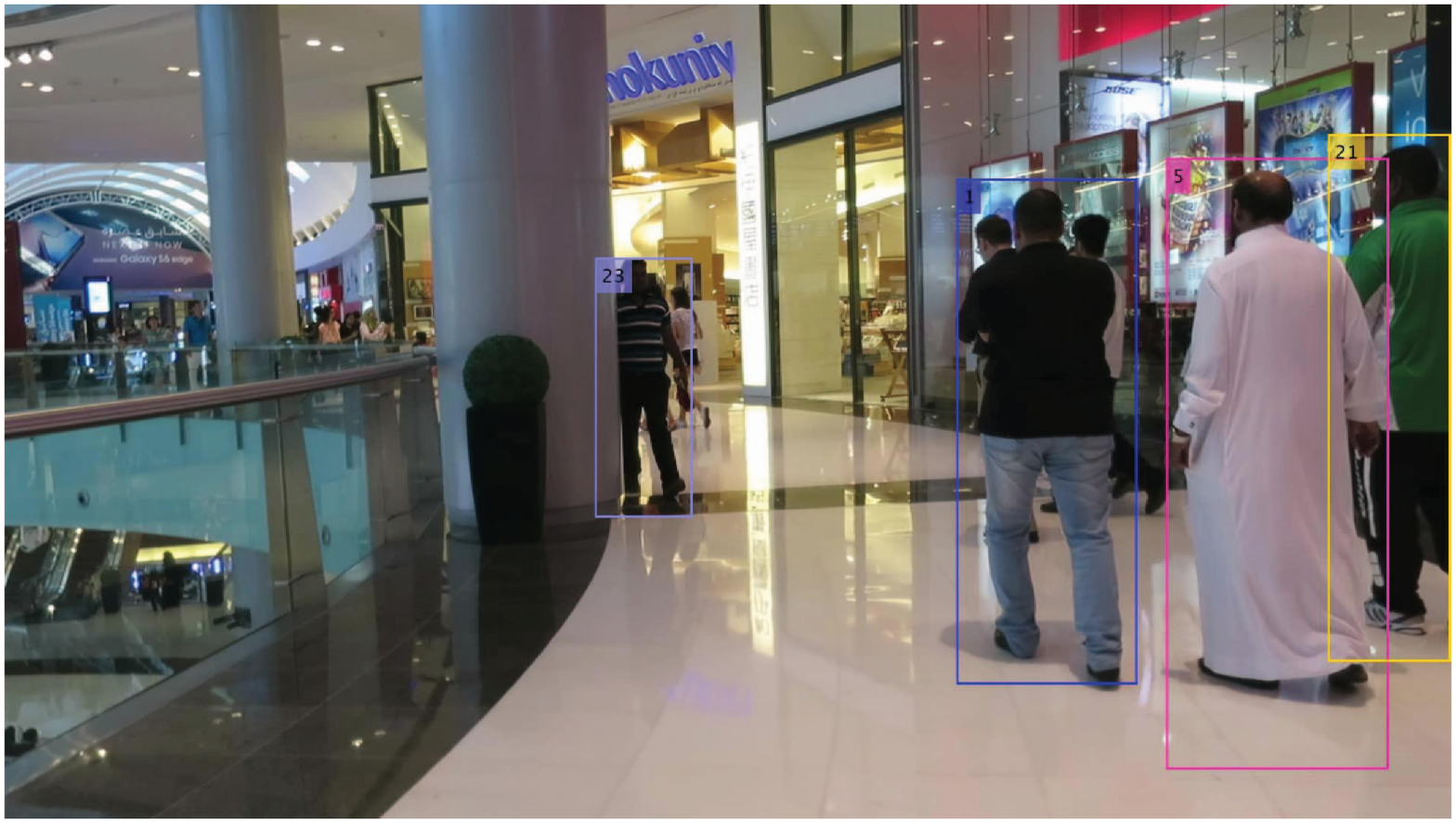}\hspace*{1em}
	\includegraphics[width=.5\linewidth]{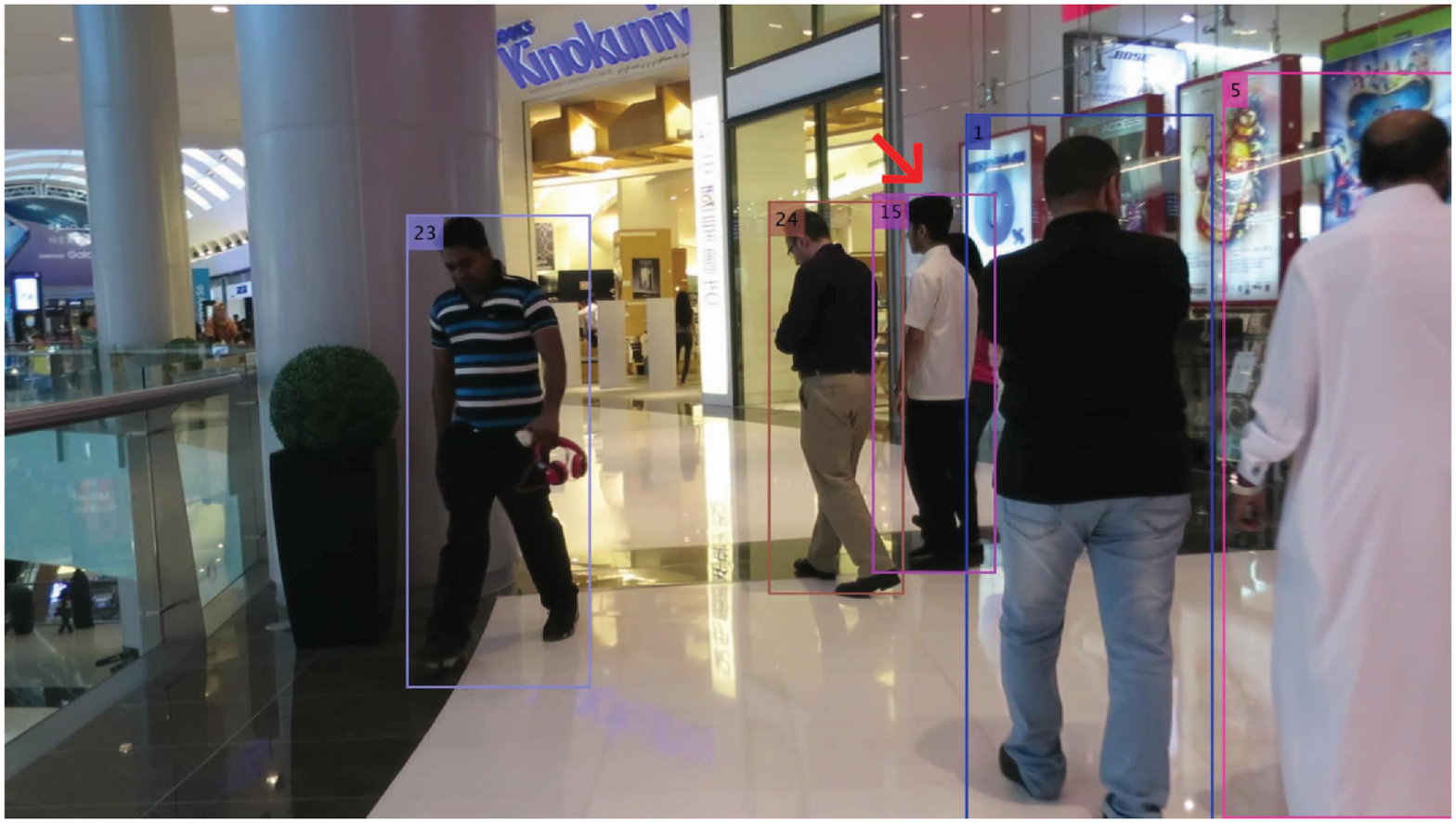}\\
	\vspace*{-2em}
	\textcolor{white}{\textbf{Frame 240}} \hspace*{35em}  \textcolor{white}{\textbf{Frame 292}} \\ 
	\vspace*{1em}
	(b)\\
	
	\caption{Illustration of qualitative results on MOT17 dataset. (a) Four sample frames of MOT17-08-FRCNN sequence. (b) Four sample frames of MOT17-12-SDP sequence. In each part, (a) and (b), there is an occluded person that is correctly labeled after reappearance.}
	\label{fig:5}
\end{figure*}

In this section the experimental results of the proposed method on well-known test datasets MOT15 \cite{leal2015motchallenge} and MOT17 \cite{milan2016mot16} are presented. In order to effectively evaluate the proposed method against state-of-the-art methods, it is necessary that identical set of metrics be used. A wide range of metrics are introduced in literature which are applied in MOT \cite{bernardin2008evaluating, li2009learning}. Based on the contributions of the proposed method some of these metrics are of more importance to be taken into consideration. As our proposed method is designed to address occlusion and miss-detection problems, and to reduce ID switches and false positives so MOTA, FP, FN, Precision and IDsw are metrics that we focus on.\\
We evaluated the proposed method on MOT17 and MOT15, two publicly available test datasets. The main difference between them is originated from the provided detections. In MOT15, just one type of detection is provided, by using ACF \cite{dollar2014fast}, while MOT17 has used three types of detectors DPM \cite{felzenszwalb2009object}, FRCNN \cite{ren2015faster}, and SDP \cite{yang2016exploit} which have returned more accurate detections. So it is expected that MOT methods show better performance on MOT17, although it contains scenes with higher track to frame ratio than MOT 15.\\
Figures \ref{fig:4} and \ref{fig:5} show the qualitative results of our proposed methods on some selected frames of MOT15 and MOT17 datasets, respectively. The selected frames contain object occlusion, and the tracking results show the correct labeling task thanks to applying occlusion handling section. Taking into account occluded tracks in each frame as a group helps to effectively overcome the occlusion issue. Note that occlusion handling section is performed using a modified version of $\delta$-GLMB filter, so the solution is also an efficient one. It also should be noted that our method performs well from the point of view of ID switch. In fact, there is a trade-off between speed and accuracy, even in the situation that the ID switch handling section of our proposed method is not a post-processing section with high computational load, and is originally a modification in the $\delta$-GLMB filter.\\
The quantitative results on MOT15 and MOT17 are shown respectively in Tables \ref{tab:1} and \ref{tab:2}. The best and second-best scores are respectively indicated by red and blue colors. As it can be seen from the tables, our methods achieves the best MOTA, MOTP, FP, Precision and IDsw in MOT15. Our method also shows the best MOTA, MOTP, and the second-best in FP, FN, Precision and IDsw in MOT17. As it was aforementioned, our contributions are based on addressing occlusion and miss-detection problems, and on reducing ID switches and false positive. As it can be seen, along with achieving the best MOTA, acceptable FP, FN and IDsw scores confirm the effectiveness of our contributions. Hz scores show we have been so focused on accuracy, although speed can be improved by optimizing the code and implementing in C++.\\
In Tables \ref{tab:1} and \ref{tab:2}, online and offline methods are tagged. Generally offline methods performs better due to their access to whole detections at once while they are not suitable for online applications. Results show that our online MOT method shows better or at least the same level of performance comparing with state-of-the-art offline MOT methods. Methods are also grouped according to be or not to be based on DNNs. Recently DNN-based methods have shown remarkable results in both detection and tracking parts of MOT task. The quantitative results indicate that our method shows better or at least the same level of performance. Methods based on using RFS theory are also grouped in the colored rows. Our method performs as the best method among those of based on RFS theory – PHD or GLMB filters. \\ 

\begin{table}[h!]
	\centering
	\caption{Illustration of quantitative results on MOT15 dataset. Our proposed method, $\textbf{MOMOT}$, is compared with other state-of-the-art online and offline methods. RFS-based methods are highlighted in colored rows and we also group the results into online and offline methods. The best and the second-best results are highlighted in red and blue colors respectively. DNN-based methods are starred.} 
	\label{tab:3}       
	\begin{adjustbox}{width=1\textwidth}
		\begin{tabular}{|c|c|c|c|c|c|c|c|c|c|c|}
			\hline
			
			\textbf{Method} & \textbf{Mode} & \textbf{MOTA} $\boldsymbol{\uparrow}$ & \textbf{MOTP} $\boldsymbol{\uparrow}$ & \textbf{MT} $\boldsymbol{\uparrow}$ & \textbf{ML} $\boldsymbol{\downarrow}$ & \textbf{FP} $\boldsymbol{\downarrow}$ & \textbf{FN} $\boldsymbol{\downarrow}$ & \textbf{Precision} $\boldsymbol{\uparrow}$ & \textbf{IDsw} $\boldsymbol{\downarrow}$ & \textbf{Hz} $\boldsymbol{\uparrow}$\\
			\hline\noalign{\smallskip}
			\hline
			CRFTrack* \cite{xiang2020end} & offline & \textcolor{blue}{$\textbf{40.0}\pm\textbf{14.5}$} & 71.9 & \textcolor{blue}{\textbf{23.0}} & \textcolor{red}{\textbf{28.6}} & 10,295 & \textcolor{red}{\textbf{25,917}} & 77.5 & 658 & 3.2\\
			
			TPM \cite{peng2020tpm} & offline & $36.2\pm12.8$ & 71.5 & 15.4 & 42.6 & \textcolor{blue}{\textbf{5,650}} & 33,102 & \textcolor{blue}{\textbf{83.4}} & 420 & 0.8\\
			
			JointMC \cite{keuper2018motion} & offline & $35.6\pm17.5$ & 71.9 & \textcolor{red}{\textbf{23.2}} & 39.3 & 10,580 & \textcolor{blue}{\textbf{28,508}} & 75.7 & 457 & 0.6\\
			\hline\noalign{\smallskip}
			\hline
			
			HybridDAT \cite{yang2017hybrid} & online & $35.0\pm13.5$ & 72.6 & 11.4 & 42.2 & 8,455 & 31,140 & 78.2 & \textcolor{blue}{\textbf{358}} & 4.6\\
			
			INARLA* \cite{wu2019instance} & online & $34.7\pm13.2$ & 70.7 & 12.5 & \textcolor{blue}{\textbf{30.0}} & 9,855 & 29,158 & 76.6 & 1,112 & 2.6\\
			
			DCCRF* \cite{zhou2018deep} & online & $33.6\pm10.9$ & 70.9 & 10.4 & 37.6 & 5,917 & 34,002 & 82.3 & 866 & 0.1\\
			
			TDAM \cite{yang2016temporal} & online & $33.0\pm00.0$ & \textcolor{blue}{\textbf{72.8}} & 13.3 & 39.1 & 10,064 & 30,617 & 75.4 & 464 & 5.9\\
			\hline
			\rowcolor{LightCyan}
			GMPHD-OGM \cite{song2019online} & online & $30.7\pm12.6$ & 71.6 & 11.5 & 38.1 & 6,502 & 35,030 & 80.2 & 1,034 & \textcolor{red}{\textbf{169.5}}\\
			\rowcolor{LightCyan}
			PHD-GSDL \cite{fu2018particle} & online & $30.5\pm14.2$ & 71.2 & 7.6 & 41.2 & 6,534 & 35,284 & 80.0 & 879 & \textcolor{blue}{\textbf{8.2}}\\
			\rowcolor{LightCyan}
			\hline
			\textbf{MOMOT(Ours)} & online & \textcolor{red}{$\textbf{40.0}\pm\textbf{08.8}$} & \textcolor{red}{\textbf{76.9}} & 6.0 & 36.9 & \textcolor{red}{\textbf{3,190}} & 33,370 & \textcolor{red}{\textbf{89.8}} & \textcolor{red}{\textbf{307}} & 0.7\\
			\hline
		\end{tabular}
	\end{adjustbox}
\end{table}

\begin{table}[h!]
	\centering
	\caption{Illustration of quantitative results on MOT17 dataset. Our proposed method, $\textbf{MOMOT}$, is compared with other state-of-the-art online and offline methods. RFS-based methods are highlighted in colored rows and we also group the results into online and offline methods. The best and the second-best results are highlighted in red and blue colors respectively. DNN-based methods are starred.}
	\label{tab:4}       
	\begin{adjustbox}{width=1\textwidth}
		\begin{tabular}{|c|c|c|c|c|c|c|c|c|c|c|}
			\hline
			
			\textbf{Method} & \textbf{Mode} & \textbf{MOTA} $\boldsymbol{\uparrow}$ & \textbf{MOTP} $\boldsymbol{\uparrow}$ & \textbf{MT} $\boldsymbol{\uparrow}$ & \textbf{ML} $\boldsymbol{\downarrow}$ & \textbf{FP} $\boldsymbol{\downarrow}$ & \textbf{FN} $\boldsymbol{\downarrow}$ & \textbf{Precision} $\boldsymbol{\uparrow}$ & \textbf{IDsw} $\boldsymbol{\downarrow}$ & \textbf{Hz} $\boldsymbol{\uparrow}$\\
			\hline\noalign{\smallskip}
			\hline
			
			TT17* \cite{zhang2020long} & offline & \textcolor{blue}{$\textbf{54.9}\pm\textbf{11.6}$} & 77.2 & \textcolor{red}{\textbf{24.4}} & 38.1 & 20,236 & \textcolor{red}{\textbf{233,295}} & 94.2 & \textcolor{red}{\textbf{1,088}} & 2.5\\
			
			TPM \cite{peng2020tpm} & offline & $54.2\pm12.2$ & 76.7 & 22.8 & 37.5 & \textcolor{red}{\textbf{13,739}} & 242,730 & \textcolor{red}{\textbf{95.9}} & 1,824 & 0.8\\
			
			CRFTrack* \cite{xiang2020end} & offline & $53.1\pm12.1$ & 76.1 & \textcolor{blue}{\textbf{24.2}} & \textcolor{red}{\textbf{30.7}} & 27,194 & 234,991 & 92.4 & 2,518 & 1.4\\
			
			eHAF17 \cite{sheng2018heterogeneous} & offline & $51.8\pm13.2$ & 77.0 & 23.4 & 37.9 & 33,212 & 236,772 & 90.8 & 1,834 & 0.7\\
			
			NOTA* \cite{chen2019aggregate} & offline & $51.3\pm11.7$ & 76.7 & 17.1 & 35.4 & 20,148 & 252,531 & 93.9 & 2,285 & 17.8\\
			
			JointMC \cite{keuper2018motion} & offline & $51.2\pm14.5$ & 75.9 & 20.9 & 37.0 & 25,937 & 247,822 & 92.4 & 1,802 & 1.8\\
			
			TLMHT \cite{sheng2018iterative} & offline & $50.6\pm12.5$ & \textcolor{blue}{\textbf{77.6}} & 17.6 & 43.4 & 22,213 & 255,030 & 93.3 & 1,407 & 2.6\\
			\hline\noalign{\smallskip}
			\hline
			
			MOTDT17* \cite{8486597} & online & $50.9\pm11.7$ & 76.6 & 17.5 & 35.7 & 24,069 & 250,768 & 92.9 & 2,474 & \textcolor{blue}{\textbf{18.3}}\\
			
			\hline
			\rowcolor{LightCyan}
			YOONKJ17 \cite{yoon2020oneshotda} & online & $51.4\pm12.9$ & 77.0 & 21.2 & 37.3 & 29,051 & 243,202 & 91.7 & 2,118 & 3.4\\
			\rowcolor{LightCyan}
			GMPHD-OGM \cite{song2019online} & online & $49.9\pm13.3$ & 77.0 & 19.7 & 38.0 & 24,024 & 255,277 & 92.8 & 3,125 & \textcolor{red}{\textbf{30.7}}\\
			\rowcolor{LightCyan}
			MTDF17 \cite{fu2019multi} & online & $49.6\pm13.7$ & 75.5 & 18.9 & \textcolor{blue}{\textbf{33.1}} & 37,124 & 241,768 & 89.7 & 5,567 & 1.2\\
			\rowcolor{LightCyan}
			\hline
			\textbf{MOMOT(Ours)} & online & \textcolor{red}{$\textbf{55.5}\pm\textbf{12.7}$} & \textcolor{red}{\textbf{77.8}} & 19.0 & 35.9 & \textcolor{blue}{\textbf{15,520}} & \textcolor{blue}{\textbf{234,235}} & \textcolor{blue}{\textbf{95.5}} & \textcolor{blue}{\textbf{1,333}} & 0.6\\
			\hline
		\end{tabular}
	\end{adjustbox}
\end{table}

\section{Conclusion and Future Works}
\label{sec:5}

This paper proposes an online MOT method which applies $\delta$-GLMB filter along with the proposed modules in order to handle occlusion and miss-detection, remove false positives (clutters) and manage MOT task by reducing ID switches and recovering missed identities. To do this, we use provided detections whose accuracies are directly effective on final tracking performance. Besides using the detections, the proposed method exploits the visual features of detections to use in track management processes. Since the $\delta$-GLMB filter has linear complexity in the number of measurements and quadratic in the number hypothesized tracks, our proposed method is the same in complexity and suitable for online applications. The proposed method is evaluated on well-known datasets, and the results confirm the effectiveness of our contributions. Our method achieves the best and the second-best in most scores not only in comparison with the other RFS-based methods, but also in comparison with state-of-the-art offline, online and DNN-based MOT methods. \\



\end{document}